\pdfoutput=1

\documentclass[11pt]{article}

\usepackage[]{ACL2023}

\usepackage{times}
\usepackage{latexsym}
\usepackage{graphicx}
\usepackage{url}
\usepackage{xcolor,colortbl}
\usepackage{soul}
\usepackage{color}
\usepackage{CJKutf8}

\usepackage[T1]{fontenc}

\usepackage[utf8]{inputenc}

\usepackage{microtype}

\usepackage{inconsolata}
\usepackage{booktabs}
\usepackage{times}
\usepackage{latexsym}
\usepackage{graphicx}
\usepackage{svg}
\usepackage{amsmath}
\usepackage{tcolorbox}\tcbuselibrary{skins}
\usepackage{subcaption}
\usepackage{multirow}
\usepackage{array}
\usepackage{makecell}
\usepackage{float}
\usepackage{booktabs}

\usepackage{array}
\newcolumntype{P}[1]{>{\centering\arraybackslash}p{#1}}

\tcbset{modelresponse/.style={
    enhanced,
    size=fbox,
    boxrule=3pt,
    arc=2mm,
    auto outer arc,
    left=10pt,
    right=10pt,
    top=10pt,
    bottom=10pt,
    fontupper=\fon{cmtt}, 
    colback=CB_gray!10,
    colframe=CB_gray!25,
    coltitle=CB_gray!20!black, 
}}

\tcbset{zeroshot/.style={
    enhanced,
    size=fbox,
    boxrule=3pt,
    arc=2mm,
    auto outer arc,
    left=5pt,
    right=5pt,
    top=5pt,
    bottom=5pt,
    fontupper=\fon{cmtt}, 
    colback=CB_pear!15,
    colframe=CB_pear!30,
    coltitle=CB_pear!25!black, 
}}

\tcbset{kshot/.style={
    enhanced,
    size=fbox,
    boxrule=3pt,
    arc=2mm,
    auto outer arc,
    left=5pt,
    right=5pt,
    top=5pt,
    bottom=5pt,
    fontupper=\fon{cmtt}, 
    colback=CB_lightCyan!15,
    colframe=CB_lightCyan!30,
    coltitle=CB_lightCyan!25!black, 
}}

\tcbset{kshotreasoning/.style={
    enhanced,
    size=fbox,
    boxrule=3pt,
    arc=2mm,
    auto outer arc,
    left=5pt,
    right=5pt,
    top=5pt,
    bottom=5pt,
    fontupper=\fon{cmtt}, 
    colback=CB_pink!15,
    colframe=CB_pink!30,
    coltitle=CB_pink!25!black, 
}}

\definecolor{CB_pear}{HTML}{BBCC33}
\definecolor{CB_pink}{HTML}{FFAABB}
\definecolor{CB_lightCyan}{HTML}{99DDFF}
\definecolor{CB_gray}{HTML}{DDDDDD}
\definecolor{pedagogyColor}{HTML}{009e73}
\definecolor{gratitudeColor}{HTML}{cc78bc}
\definecolor{personalColor}{HTML}{ca9161}

\colorlet{pedagogyTextColor}{pedagogyColor!40}
\colorlet{gratitudeTextColor}{gratitudeColor!40}
\colorlet{personalTextColor}{personalColor!40}

\DeclareRobustCommand{\hlpedagogy}[1]{{\sethlcolor{pedagogyTextColor}\hl{#1}}}
\DeclareRobustCommand{\hlgratitude}[1]{{\sethlcolor{gratitudeTextColor}\hl{#1}}}
\DeclareRobustCommand{\hlpersonal}[1]{{\sethlcolor{personalTextColor}\hl{#1}}}

\newcommand{\fon}[1]{\fontfamily{#1}\selectfont}

\definecolor{CB_orange}{HTML}{EE8866}
\definecolor{CB_mint}{HTML}{44BB99}

\newcommand\trueLabel{\cellcolor{CB_mint!30}$\mathbf{1}$}
\newcommand\falseLabel[1]{\cellcolor{CB_orange!30}$\mathbf{0}$}

\def\datasetname{\textsc{Sight}}
\def\numComments{15,784}
\def\numManualComments{280}
\def\numLectures{288}
\def\numPlaylists{10}
\def\numLabels{9}

\def\humanAgreement{\texttt{human}}
\def\humanModelZeroShotAgreement{\texttt{0-shot}}
\def\humanModelThreeShotAgreement{\texttt{3-shot}}
\def\humanModelThreeShotReasoningAgreement{\texttt{3-shot-R}}

\usepackage[symbol]{footmisc}

\def\general{\texttt{general}}
\def\setup{\texttt{setup}}
\def\pedagogy{\texttt{pedagogy}}
\def\confusion{\texttt{confusion}}
\def\gratitude{\texttt{gratitude}}
\def\clarification{\texttt{clarification}}

\def\nonenglish{\texttt{nonenglish}}
\def\personal{\texttt{personal}}
\def\na{\texttt{na}}

\title{\datasetname: A Large Annotated Dataset on \\ 
\underline{S}tudent \underline{I}nsights \underline{G}athered from \underline{H}igher Education \underline{T}ranscripts}

\author{
Rose E. Wang\footnote{} \\ \texttt{rewang@cs.stanford.edu} \And 
 Pawan Wirawarn$^*$ \\ \texttt{pawanw@stanford.edu}  \AND
Noah Goodman \\ \texttt{ngoodman@stanford.edu} \And 
Dorottya Demszky \\ \texttt{ddemszky@stanford.edu}\\ \AND
Stanford University
}

\begin{document}
\maketitle
\begin{abstract}
Lectures are a learning experience for both students and teachers.
Students learn from teachers about the subject material, while teachers learn from students about how to refine their instruction.
However, online student feedback is unstructured and abundant, making it challenging for teachers to learn and improve. We take a step towards tackling this challenge.
First, we contribute a dataset for studying this problem: \datasetname{} is a large dataset of \numLectures{} math lecture transcripts and \numComments{} comments collected from the Massachusetts Institute of Technology OpenCourseWare (MIT OCW) YouTube channel.
Second, we develop a rubric for categorizing feedback types using qualitative analysis. 
Qualitative analysis methods are powerful in uncovering domain-specific insights, however they are costly to apply to large data sources.
To overcome this challenge, we propose a set of best practices for using large language models (LLMs) to cheaply classify the comments at scale.
We observe a striking correlation between the model's and humans' annotation: 
Categories with consistent human annotations (>$0.9$ inter-rater reliability, IRR) also display higher human-model agreement (>$0.7$), while categories with less consistent human annotations ($0.7$-$0.8$ IRR) correspondingly demonstrate lower human-model agreement ($0.3$-$0.5$).
These techniques uncover useful student feedback from thousands of comments, costing around $\$0.002$ per comment.
We conclude by discussing exciting future directions on using online student feedback and improving automated annotation techniques for qualitative research.\footnote{
Equal contributions. \\
\datasetname{} is intended for research purposes only to promote better understanding of effective pedagogy and student feedback.
We follow MIT's Creative Commons License. 
The dataset should not be used for commercial purposes. 
We include an elaborate discussion about limitations of our dataset in Section~\ref{sec:limitations} and about the ethical use of the data in the Ethics Statement Section. 
The code and data are open-sourced here: \url{https://github.com/rosewang2008/sight}.
}

\end{abstract}

\begin{figure*}[ht]
    \centering
    \includegraphics[width=\linewidth]{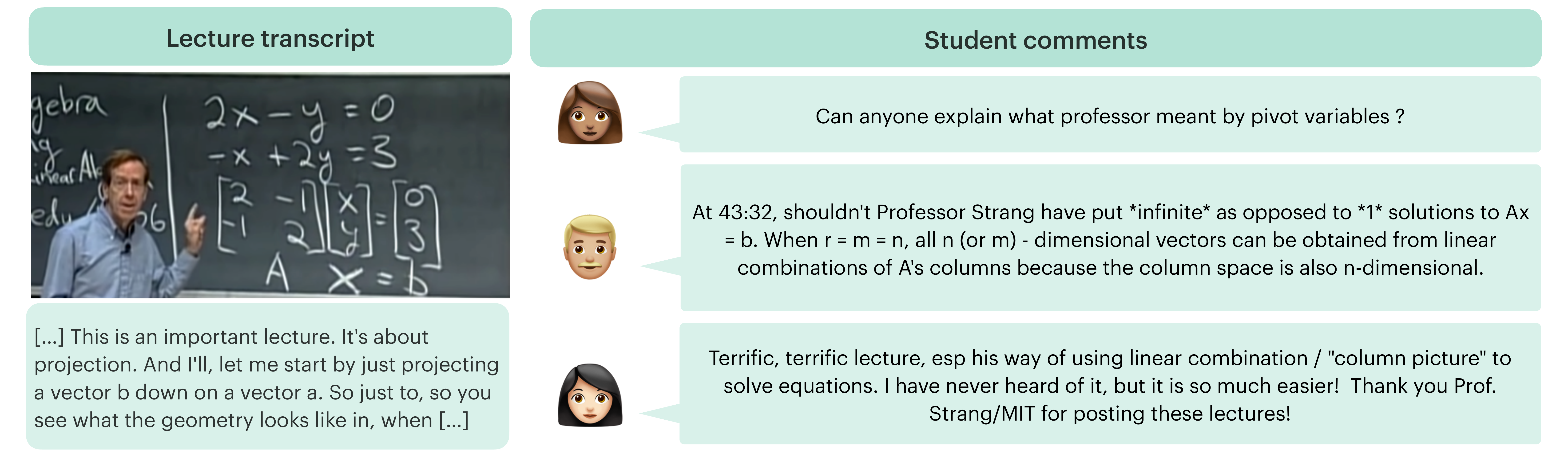}
    \caption{A peek into \datasetname: Every lecture is associated with student comments. \datasetname{} contains \numPlaylists{} courses, \numLectures{} lectures, and \numComments{} comments. The comments are labeled using our coding rubric that isolates different types of student feedback.}
    \label{fig:main_figure}
    \vspace{-1.5em}
\end{figure*}

\section{Introduction}
Lectures are a learning experience for both students and teachers.
Students learn from teachers about the subject material.
Teachers also learn from students about how to improve their instruction \citep{learning2011measuring, pianta2008classroom, evans1978clarity, hativa1998lack}. 
However, in the online education setting, student feedback is both abundant and unstructured.
This makes it challenging for teachers with online content to synthesize and learn from available feedback.

To take a step towards tackling this challenge, we contribute \datasetname{} (\underline{S}tudent \underline{I}nsights \underline{G}athered from \underline{H}igher Education \underline{T}ranscripts), a large dataset of \numLectures{} math lecture transcripts and \numComments{} comments collected from the Massachusetts Institute of Technology OpenCourseWare (MIT OCW) YouTube channel.
MIT OCW is a popular YouTube channel that offers a collection of lecture content from real MIT courses. 
Their courses gather up to thousands of student comments \citep{mit_2020, mit_ocw,breslow2013studying}, in which users express a range of feedback from excitement about the pedagogy to confusion about the course content. 
The dataset is a rich source of data for studying the relationship between teaching content and student commentary. 

Second, we develop a rubric for categorizing different kinds of student feedback in the YouTube comments using a qualitative analysis approach.
Qualitative analysis involves iteratively examining the data and accounting for the context \citep{corbin1990basics, erickson1985qualitative,bauer2000qualitative}.
For example, we examine the student comments for useful feedback categories while accounting for the \textit{online} context of the instruction.
Our rubric includes \numLabels{} categories of student YouTube comments, spanning from general feedback useful for encouraging instructors (e.g., ``Amazing lectures!'') to specific comments on the pedagogy or technical content.

While qualitative analysis methods are effective in uncovering domain-specific insights, applying these methods to large sources of data is challenging \citep{erickson1985qualitative,corbin1990grounded, bauer2000qualitative,o2020intercoder}.
Scaling annotation effectively is crucial for sifting through large amounts of unstructured data (e.g., the \numComments{} comments in \datasetname{}) and uncovering relevant student feedback. 
However, qualitative methodologies often require domain-expertise. 
This limits the pool of analysts, which makes it expensive to find this expertise or means that only a small sample of the data can be analyzed \citep{harrison2019investigating, lee2017making}.
Additionally, the qualitative analysis process is time-consuming because it allows for the annotation rubric to be adapted and accommodate new categories.
This means that data has to be re-annotated frequently and must be analyzed flexibly. 
Therefore, our third contribution is proposing a set of best practices for using pretrained large language models (LLMs)---specifically ChatGPT \citep{openai_chatgpt_2023}---to \textit{cheaply, quickly} and \textit{flexibly} annotate data at scale. 
We explore different prompting approaches (e.g., zero-shot, k-shot, and reasoning).

We analyze the quality of the model annotation and the diversity of user feedback.
Categories with consistent human annotations (>$0.9$ inter-rater reliability, IRR) also display higher human-model agreement (>$0.7$), while categories with less consistent human annotations ($0.7$-$0.8$) correspondingly demonstrate lower human-model agreement ($0.3$-$0.5$).
Albeit imperfect, annotating with ChatGPT allows researchers to explore their entire dataset in a fast, cost-effective way. 
For example, we are able to sift through \numComments{} comments  and identify those related to student confusion and lecture pedagogy in a few hours, all under $\$0.002$ per comment.
These comments can be invaluable for instructors looking to improve their lecture content. 

In summary, we make the following contributions in this paper:
\begin{enumerate}
    \item We create \datasetname{}, a dataset of \numLectures{} lecture transcripts from MIT OpenCourseWare (OCW) mathematics courses and of \numComments{} annotated user comments.  
    \item We develop an annotation rubric of feedback types found in YouTube comments using a qualitative analysis approach. 
    \item We release a set of best practices for using LLMs with qualitative coding rubrics for scaling annotation.
    \item We analyze the quality of the annotation and the diverse types of student feedback uncovered via our automated annotation procedure.
\end{enumerate}

\section{Related Work}

\subsection{YouTube as an Educational Platform}

YouTube is an online platform with a vast collection of educational videos, such as from MIT OCW \citep{mit_ocw}. 
Due to its popularity and large volume of high-quality education content, YouTube is an important platform for providing educational resources.
For example, prior work in education have studied how YouTube provides educational video content to fields like medicine \citep{curran2020youtube}, support multi-modal learning through video and lecture slides \citep{lee2022multimodal}, or allows for informative discussions in the comment section \citep{dubovi2020empirical,lee2017making}.
Prior works have not yet studied how YouTube comments can serve as feedback for instructors who host online content.

\subsection{Student Feedback}
Course evaluations by students are the cornerstone for providing feedback to instructors at higher educational institutions \citep{hammonds2017student,marsh1997making}.
However, internal course evaluations receive limited student responses, are administered infrequently, and suffer from recency bias (e.g., surveys are typically administered after final examinations) \citep{cohen1981student,greenwald1997grading, kim2023high}.
Integrating traditional evaluations with informal evaluations from online platforms, like MOOCs and YouTube, can expand the sources of feedback.

Another challenge with course evaluations is that they are unstructured: Student evaluations can encompass a wide range of topics, from technical issues to personal opinions. 
This is an even more prominent issue for YouTube where videos receive a lot of spam comments. 
While unsupervised natural language processing (NLP) methods, such as topic modeling, have been applied to survey data to extract themes \citep{hujala2020improving}, they struggle to identify specific information.
Alternatively, classifiers can be trained for specific domains, e.g., sentiment classifiers for measuring class mood  \citep{hynninen2019distinguishing,baddam2019student,gottipati2018text, alhija2009student,azab2016analysing}, however they are time-consuming to train, especially if the rubrics are modified over the course of analysis.
Finally, although qualitative analysis offers powerful insights, it is typically limited to small data samples and is challenging to scale
\citep{asselin2011learning,brook2011affordances,lee2017making}.

\subsection{LLMs for Qualitative Analysis}
Recent advances in NLP have resulted in the development of sophisticated pretrained LLMs like ChatGPT \citep{openai_chatgpt_2023}. 
These models are appealing because they are able to generalize to many domains and follow instructions easily \citep{brown2020language}.
We believe these characteristics are particularly appealing for researchers who use qualitative analysis methods and want to explore their dataset fully. 
Recent works have explored using ChatGPT for annotation on \textit{existing} datasets and benchmarks \citep{kuzman2023chatgpt,ziems2023css, he2023annollm,gilardi2023chatgpt}.
Our work explores applying LLMs to scale annotation on a novel rubric we've designed for our research purposes.

\section{\datasetname{} \label{sec:dataset}}

\begin{table}
\centering
\begin{tabular}{c|c}
\toprule
\textbf{Number of lecture series} & \numPlaylists{} \\
\textbf{Number of lecture transcripts} & \numLectures{} \\
\textbf{Number of comments} & \numComments{} \\
\textbf{Number of labels} (Section~\ref{sec:comment_rubric}) & \numLabels{} \\
\bottomrule
\end{tabular}
\caption{Summary statistics for the \datasetname{}. We use the labels developed from the coding rubric described in Section~\ref{sec:comment_rubric} to annotate all the comments in the dataset.}
\label{tab:summary_statistics}
\vspace{-1em}
\end{table}

\begin{table*}[t]
\centering\small 
\begin{tabular}{p{2.0cm}|p{10.2cm}|P{1.0cm}|P{1.0cm}}
\toprule
\small
\textbf{Category} & \textbf{Example comment} & \textbf{\%} & \textbf{\#}\\

\midrule
\general{} & Best video I have watched so far, I was with him all the way and my concentration never dipped. & $28.37$\%  & $82$ \\

\midrule
\confusion{} & 34:43 why "directional second derivative" would not give us a clue of whether it is a min or max? I thought it is a promising way. hmmm. & $20.76$\% & $60$ \\

\midrule
\pedagogy{} & From this lecture, I really understand Positive Definite Matrices and  Minima thanks  to Dr. Gilbert Strang. The examples really help me to fully comprehend this important subject. & $7.27$\% & $21$\\

\midrule
\setup{} & Oh.. my god.. the board and chalk are phenomenal..! & $3.81$\% & $11$\\

\midrule
\personal{} & sweet, did this like a term and a half ago in higshcool. aced the test for it too :D
gosh calculus is awesome! & $9.00$\% & $26$ \\

\midrule
\clarification{} & @[USERNAME] Actually, if a constant k=1/1m is used, then in the final formula for V you will end up with subtracting m\^{}1 from m\^{}2 which is apparently not correct. & $2.42$\% & $7$ \\

\midrule
\gratitude{} & Thank you very much! Amazing lectures! & $13.49$\% & $39$ \\

\midrule

\nonenglish{} & Tłumaczenie na polski wymiata & $6.57$\%& $19$ \\

\midrule
\na{} & sounds drunk on 0.5 speed & $42.21$\% & $123$ \\

\bottomrule

\end{tabular}
\caption{
Example comments for each comment annotation category.
The category percentage of the sample dataset is reported in the column \%.
Note, a comment can be labeled with multiple categories so the percentages do not add up to 100\%.
The number of comments in the sample dataset labelled with that category by at least one of the annotators is reported in the column \#.
}
\label{tab:example_comments}
\end{table*}

This section details the dataset contents and data collection procedure.
Table \ref{tab:summary_statistics} summarizes the dataset statistics.

\subsection{Lecture Transcripts}
Our work focuses on math lectures from MIT OCW. 
We use all of the math course playlists listed on MIT OCW as of date, and all of the videos belonging to those playlists. 
This altogether gives \numPlaylists{} playlists with \numLectures{} videos.
Each playlist has up to 35 lecture videos.
These playlists range from general mathematics courses on calculus and linear algebra to more advanced topics like graph theory and functional analysis. 
For the full list, please refer to Appendix~\ref{sec:app_dataset}.

We use the Google YouTube API to extract the video identification numbers within each playlist, and the YouTube Data API V3 to collect the audio from each video. 
To transcribe the video audio to text, we use OpenAI's Whisper \texttt{large-v2} model \citep{radford2022robust}. 
We manually check the quality of some of the lecture transcripts and find them to be faithful to what is said in the lectures.
Our dataset tracks each lecture's video ID, video title, playlist ID, and transcription model used.

\begin{figure*}[t]
    \centering 
    \begin{tcolorbox}[
    zeroshot,
    title={\small \textbf{Zero-shot prompting for \pedagogy{} category}},
    ]
    \small 
    Consider a YouTube comment from the math MIT OCW video below:\\
    
Playlist name: \{playlistName\}\\
Video name: \{videoName\}\\
Comment: \{comment\} \\

If the statement below is true, please respond "true"; otherwise, please respond "false": 

The comment mentions the teacher’s instructional method, which includes but is not limited to the use of examples, applications, worked out problems, proofs, visualizations, elaboration, and analogies.

    \end{tcolorbox}
    \vspace{-1em}
    \caption{
    The zero-shot prompt for the \pedagogy{} category.
    \label{fig:example_zeroshot_prompt}
    }
\end{figure*}

\subsection{Lecture Comments}

We use the Google YouTube API to collect a total of \numComments{} user comments from each lecture videos.
We do not track the user ID of the comment.
If the comment mentions another user with ``@'', we anonymize the username by replacing it with ``[USERNAME]''.
All comments are top-level comments, not replies to comments.
This means that if a comment belongs in a thread of another comment, it is not included in our dataset. 
Each course playlist varies in the number of comments; Appendix~\ref{sec:app_dataset} reports the comment statistics.
We annotate the comments according to a coding rubric we develop to better understand how users engage with the instruction and lecture content. 
This rubric is detailed in the next section, Section~\ref{sec:comment_rubric}.

\section{Feedback Rubric \label{sec:comment_rubric}}

We develop a rubric that catalogs different types of student feedback found in \datasetname.
This rubric is used for annotating the comments at scale as well (rf. Section~\ref{sec:scaling_annotation}).
This section details how we developed this rubric.

\subsection{Rubric Development}

When creating the taxonomy, we seek to jointly maximize the following objectives.
\begin{itemize}
    \item \textit{Provide coverage of feedback expressed in our data.} We uncover categories starting from the data and manually label a subset of the data; this is a part of a qualitative research methodology known as the grounded theory approach \citep{corbin1990grounded}.
    \item \textit{Provide coverage of feedback types in the literature.} After we developed a set of categories directly from the data, we consult prior work on course evaluations to incorporate potentially missing themes. Specifically, we used \citet{gravestock2008student, chen2003student, zabaleta2007use, kim2023high} as additional sources.
    \item \textit{Be specific about what the feedback is about.} We want to make the feedback categories targeted, enabling instructors to easily understand areas for improvement.
\end{itemize}

Our process for developing the rubric follows the procedure outlined in \citet{o2020intercoder,seidel2015beyond, corbin1990grounded}: 
Two authors read a subset of randomly selected comments and developed the initial categories collaboratively. The categories were then adapted to be specific and iterated until both authors agreed that the categories sufficiently covered the comments.

\subsection{Rubric Categories}

The final feedback categories in our rubric are detailed below. 
Examples of each category are shown in Table~\ref{tab:example_comments}.

\paragraph{General:} The comment expresses a general/big-picture opinion about the video's content and/or about the teaching/professional characteristics of the instructor. 
For example, ``Amazing!!!'' or ``Great teacher.'' would be marked as \general.

\paragraph{Confusion:} The comment asks a math-related question, expresses math-related confusion, and/or points out a math-related mistake in the video.

\paragraph{Pedagogy:} The comment mentions an instructional method. Instructional methods include the use of examples, applications, worked out problems, proofs, visualizations, elaboration, and analogies.

\paragraph{Teaching setup:} The comment describes or mentions the lecture’s teaching setup. 
Teaching setup includes the chalk, chalkboard, microphone or audio-related aspects, and camera or camera-related aspects (e.g., angle).

\paragraph{Personal experience:} The comment mentions the user’s personal experience or context with respect to the lecture. 
Personal experience or context includes the user’s own math learning or teaching experiences.

\paragraph{Clarification:} The comment clarifies someone's math-related misunderstanding or elaborates content from the video, and the comment includes an `@` that is immediately followed by a username.

\paragraph{Gratitude:} The comment contains the word ``thanks'' or ``thank''.


\paragraph{Non-English comment:} The comment is not in English.\footnote{Because the lectures are conducted in English and the authors feel most comfortable English, we make the distinction between English and non-English comments.}

\paragraph{N/A:} The comment expresses a joke or is a troll comment, and/or the comment says something that is hard to connect to the video content, and/or the comment does not fall into any of the categories above.

\subsection{Annotation of Sample Dataset \label{sec:sample_dataset}}
We have two annotators (co-authors) annotate a sample dataset of \numManualComments{} comments based on the rubric descriptions provided above.
The annotators are asked to select all categories that applied.
Table~\ref{tab:example_comments} reports the category percentage in the sample dataset. 
Appendix~\ref{app:human_annotation} includes an image of the annotation interface.

\section{Scaling Annotation \label{sec:scaling_annotation}}

This section details how we scale annotations using our rubric and LLMs.
Scaling annotation is crucial for sifting through large amounts of unstructured data (e.g., the \numComments{} comments in \datasetname{}) and uncovering relevant student feedback. 
By using LLMs, we can cheaply and quickly classify comments, without the need for expensive human annotation.

\subsection{Model}
For scaling annotation, we use GPT-3.5 (\texttt{gpt-3.5-turbo}) through the OpenAI API \citep{openai_chatgpt_2023}.
Although alternative models can also be used, such as \texttt{text-davinci-003} from the original InstructGPT model family \citep{ouyang2022training}, or open-sourced models like Flan-T5 \citep{chung2022scaling}, GPT-3.5 is  cheap, effective, and generally accessible for researchers without GPU support.

\subsection{Prompting Methods \label{prompting_methods}}
This section discusses the prompting strategies used for scaling annotation.
We also experiment with other approaches, but report the most effective approaches in the main text. 
We detail our prior attempts and best practices in Appendix~\ref{app:scaling_annotation}, which we believe to be highly instructive for researchers applying this methodology to other settings. 

Each comment is annotated as a binary classification task per category, i.e., does this category apply to this comment? 
We found that comments oftentimes contained multiple types of feedback. 
For example, a comment like ``His \hlpedagogy{teaching style seems casual and intuitive}. \hlpersonal{I go to a small public college} and the course is much more \hlpedagogy{formal and proof driven}. These lectures are a great addition to (as well as a nice break from)  formal proofs. \hlgratitude{Thanks} MIT!'' includes feedback tied to the \hlpedagogy{\pedagogy}, \hlpersonal{\personal}, and \hlgratitude{\gratitude{}} categories.

\begin{table*}[t]
    \centering
    \small
    \begin{tabular}{c c c c c c c c c c}
    \toprule
    \bf IRR  & \texttt{gen.} &  \texttt{conf.} & \texttt{peda.} & \texttt{set.} & \texttt{pers.} & \texttt{clar.} & \texttt{gra.} & \texttt{noneng.} & \na{} \\ 
    \midrule
    \rowcolor{lightgray} \bf \texttt{human} & $0.75$ & $0.91$ & $0.79$ & $0.95$ & $0.74$ & $0.83$ & $0.92$ & $0.94$ & $0.74$\\ 
    \midrule
    \bf \texttt{0-shot} & $0.48$ & $0.72$ & $ 0.35$ & $\bf 0.85$ & $0.32$ & $\bf 0.48$ & $0.92$ & $\bf 0.87$ & $\bf 0.65$\\ 
    \bf \texttt{3-shot} & $0.50$ & $0.69$ & $0.52$ & $0.75$ & $\bf 0.57$ & $0.16$ & $0.85$ & $0.64$&  $0.50$\\ 
    \bf \texttt{3-shot-R} & $\bf 0.52$ & $\bf 0.76$ & $\bf 0.57$ & $\bf 0.85$ & $0.37$ & $0.32$ & $\bf 0.93$ & $0.50$ & $0.47$ \\ 
    \bottomrule
    \end{tabular}
    \caption{
    Cohen's kappa scores for measuring inter-rater reliability (IRR) within humans (\humanAgreement) and within human-model pairs across the rubric categories (abbreviated in the table).
    We bold the best human-model strategy within each category.
    The \humanAgreement{} IRR is used as a reference score.
    It is in the highlighted row and always reaches substantial to perfect agreement (at least $0.70$).
    The other rows measure the average human-model IRR  when the model is prompted 0-shot (\humanModelZeroShotAgreement), 3-shot (\humanModelThreeShotAgreement), or 3-shot with reasoning (\humanModelThreeShotReasoningAgreement).
    }
    \label{tab:correlation_scores_class}
\end{table*}

\begin{table*}[ht]
\centering\small 
\begin{tabular}{P{0.5cm} | p{2.0cm} | p{10cm} | P{0.5cm} | P{0.5cm} }
\toprule

\bf \# & \bf Category & \bf Comment & \centering \bf H & \bf M \\

\midrule 

A & \pedagogy{} & This guy is great.  I studied engineering at a university less prestigious than MIT, and I remember professors refusing to explain their algebra steps.  They were like "you should know this already". &  \trueLabel & \falseLabel{} \\
\midrule
B & \personal{} & Wish this guy taught me Math 293 and 294 at Cornell. My guy could barely speak English, let alone explain what we were trying to accomplish. I understood that if we wanted eigenvectors perpendicular to x we'd get lift relative to flow...but this guy would have made the math a bit simpler. &  \trueLabel & \falseLabel{} \\
\midrule
C & \pedagogy{} & 41:53 These are questions that should be asked in recitation, not in lecture. &  \falseLabel{} & \trueLabel{} \\
\midrule
D & \personal{} & why is iteration in newtons done..i cant understand the logic behind this &  \falseLabel{} & \trueLabel{} \\
\bottomrule

\end{tabular}
\caption{
Error analysis on \pedagogy{} and \personal{}, the two lowest agreement categories on the zero shot setting (\humanModelZeroShotAgreement).
The \textbf{H} column is the category label that both humans assigned the comment to, and the \textbf{M} column is the label that the model assigned the comment to. 
$1$ indicates that the annotator believes the category \textit{does} apply to the comment, whereas $0$ is where the category is presumed \textit{not} to apply.
\label{tab:error_analysis}
}
\end{table*}

\begin{table*}[ht]
\centering\small 
\begin{tabular}{P{2.2cm} | p{13.1cm}}
\toprule

\textbf{Subcategories} & \textbf{Comments labeled as \confusion{} }\\

\midrule
Conceptual & Can anyone explain what professor meant by pivot variables ? \\

\midrule
Conceptual &  Can anyone help me understand, why the professor keep saying at 19:01 that we can't solve 4 equation with 3 unknowns?
\\

\midrule
Potential mistake & i think the explanation of the first queston was a little bit wrong it seems. because he wrote the equation  to diagonalize the matxix P even though it does not have 3 independent eigen vectors. \\

\midrule
Potential mistake & Anyone understand the equation at 32:15? I think x\_free should be above x\_pivot? \\

\midrule
Resources &  What is good homework to test if we clearly understand this lecture? Is there such corresponding homework?
\\

\midrule
Resources &  Does anyone know which lecture he derive the general equation for a determinant? Would be a massive help thanks! \\

\bottomrule

\end{tabular}
\caption{Example comments in the \confusion{} category.}
\label{tab:confusion_comments_diversity}
\vspace{-1.5em}
\end{table*}

\paragraph{Zero-shot prompting.}
Zero-shot prompting directly asks the model to label the category.
Following prior work \citep{child2019generating,ziems2023css}, we first provide the context of the comment and the comment itself, \textit{then} provide instructions on the labelling task. 
The context of the comment includes a mention to MIT OCW, the playlist name and video name. 
The instructions include a description of the category, and prompts the model to respond with ``true'' or ``false'' for whether the category applies to the comment. 
Figure~\ref{fig:example_zeroshot_prompt} shows an example of a zero-shot prompt.
The zero-shot setting is the most similar to the human annotation setup.

\paragraph{K-shot prompting.}
K-shot prompting provides examples of the annotation.
It first includes the instructions at the top of the prompt, then $k$ examples that include the context, the comment, and the label.
Our work uses 3-shot examples.
We did not find any benefits including more than 3 examples. 
The instructions are moved to the top to avoid repeating the instructions after every example.
Due to space constraints, we include our k-shot prompts in Appendix~\ref{app:all_prompts}.

\paragraph{K-shot prompting with reasoning.}
K-shot prompting with reasoning is similar to k-shot prompting, but additionally provides a reasoning for the label. 
The reasoning comes after the comment, but before the label. 
Due to space constraints, we include our k-shot reasoning prompts in Appendix~\ref{app:all_prompts}.

\newcommand*{\escape}[1]{\texttt{\textbackslash#1}}

\subsection{Evaluation}

We aim to measure the effectiveness of ChatGPT in scaling the annotation process and providing instructors with useful feedback. Our evaluations are centered around the following research questions.

\begin{description}
    \item[RQ1:] \emph{How does the zero-shot approach with ChatGPT compare to human annotations across categories?} We investigate this question by measuring the IRR between the human annotations and the zero-shot ChatGPT annotations.
    \item[RQ2:] \emph{How does the incorporation of additional information, such as k-shot examples and reasoning, affect the model's annotation?} We investigate this question by measuring the IRR between the human annotations and the k-shot and k-shot with reasoning ChatGPT annotations.
    \item[RQ3:] \emph{What are some examples of useful feedback in the scaled annotated dataset?} We investigate this by performing a qualitative analysis (grounded theory approach) on the comments annotated with the \confusion{} category.
\end{description}

\begin{figure}[t]
    \centering
    \includegraphics[width=\linewidth]{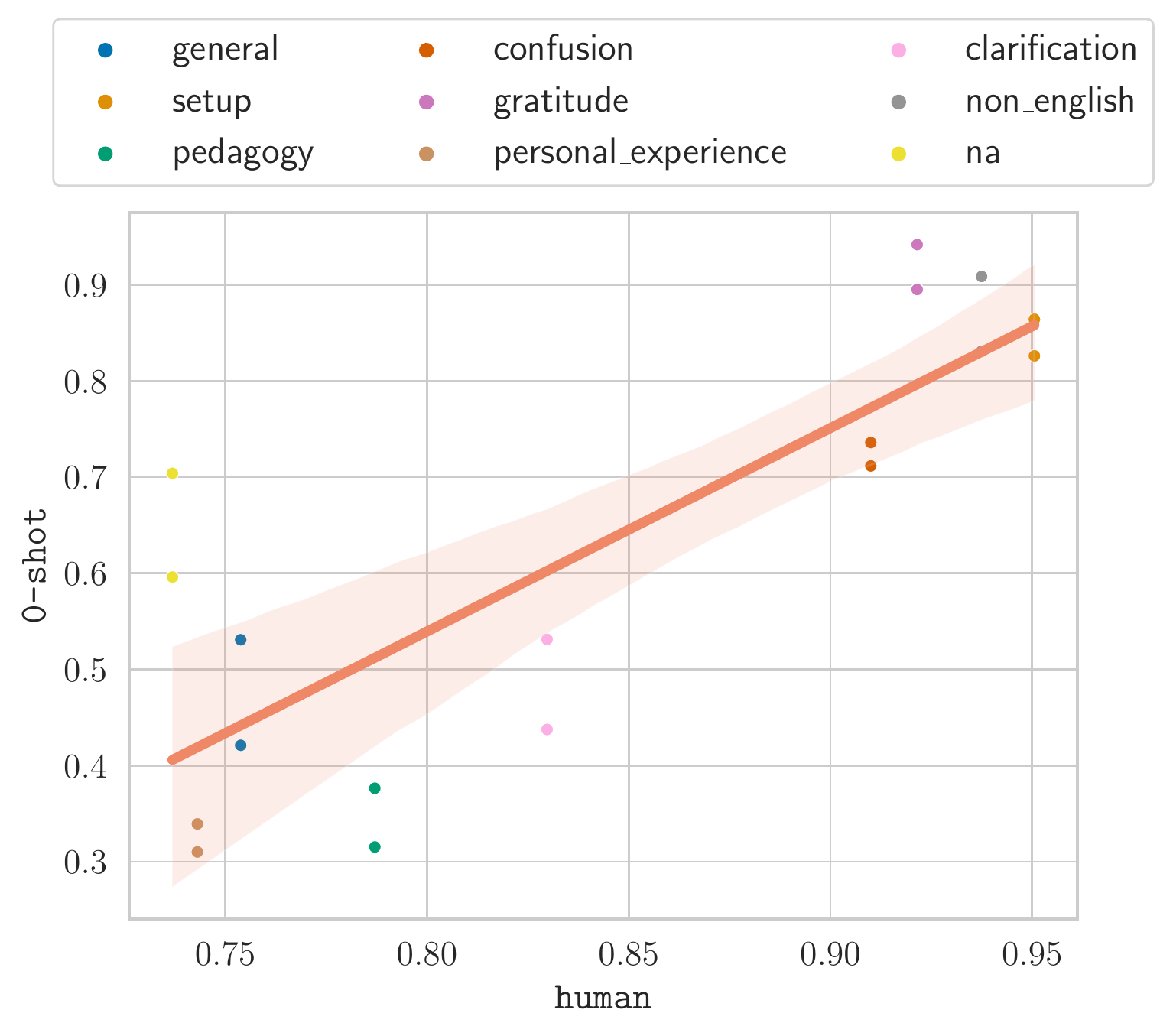}
    \vspace{-1em}
    \caption{Human inter-annotator agreement (\humanAgreement) vs. human-model inter-annotator agreement (\humanModelZeroShotAgreement).
    The agreement scores are color-coded by category. 
    }
    \label{fig:agreement_regression}
    \vspace{-1.0em}
\end{figure}
\section{Results \label{sec:analysis}}

\paragraph{RQ1.} 
We compare ChatGPT's zero-shot annotation to human annotations on the sample dataset described in Section~\ref{sec:sample_dataset}.
We compute Cohen's kappa to measure IRR within categories.
Table~\ref{tab:correlation_scores_class} reports the human IRR as \humanAgreement{} and the average human-model IRR on the zero-shot setting as \humanModelZeroShotAgreement{}.
The human-model agreement never surpasses the human agreement scores.
The human-model agreement also varies a lot across categories.
For example, the model has fair agreement with the human annotators on \pedagogy{} and \personal{} ($\sim0.30$) and perfect agreement on \gratitude{} ($0.92$).
To better understand the model's failure modes, we manually inspect the model's mislabeled comments.
Table~\ref{tab:error_analysis} shows comments that are representative of common failure modes on \pedagogy{} and \personal.
The model seems to miss subtle references to either category, such as ``explaining their algebra steps'' for category \pedagogy.
The model also tended to mislabel student questions as examples of \pedagogy{} and \personal{}.
\textbf{Categories that require more interpretation seem to be more difficult for the model to annotate in agreement with humans.}

To further investigate this, we plot \humanAgreement{} against \humanModelZeroShotAgreement{} in Figure~\ref{fig:agreement_regression}.
Strikingly, we observe a correlation between the model's annotations and the humans' annotations: 
Categories exhibiting greater consistency among human annotators (>$0.9$ IRR) also display higher agreement between humans and the model (>$0.7$), while categories with less consistent human annotations ($0.7$-$0.8$) correspondingly demonstrate lower levels of human-model agreement ($0.3$-$0.5$).
\textbf{Our findings suggest that the model's annotations reflect the variability observed in human opinions}, providing a complementary perspective to recent works such as \citet{he2023annollm}, which report models outperforming humans in annotation tasks. 
Our results suggest that this superior performance may not always hold, as the model's annotation accuracy appears to be influenced by the level of human agreement. 
Appendix~\ref{sec:category_distribution} includes additional plots of the category distributions across different annotators. 

\paragraph{RQ2.}
The previous section on RQ1's zero-shot performance indicates that the model poorly annotates categories that involve more qualitative interpretation.
Each category has seems to have common failure modes. 
This section explores potential remedies that provide \textit{more} information to the model:
We provide three examples with annotated labels for each category in the prompt. 
These examples were selected based on the errors the model made in the zero-shot setting. 
We also test providing the same three examples with human-written reasoning as to why those examples are annotated with such a label.

Table~\ref{tab:correlation_scores_class} reports the human-model agreement scores on the 3-shot (\humanModelThreeShotAgreement) and 3-shot with reasoning setting (\humanModelThreeShotReasoningAgreement). 
\textbf{The effect of the auxiliary information varies  across the categories}:
Some categories benefit from the examples and reasoning such as \pedagogy{} which does better on \humanModelThreeShotAgreement{} by +$0.17$ and on \humanModelThreeShotReasoningAgreement{} by +$0.22$ compared to \humanModelZeroShotAgreement.
However, other categories exhibit consistently worse agreement with more information, such as \clarification{} which does worse on \humanModelThreeShotAgreement{} by -$0.32$ and on \humanModelThreeShotReasoningAgreement{} by -$0.16$ compared to \humanModelZeroShotAgreement. 
We also experiment with selectively picking the examples and tuning the reasoning, but those attempts did not result in better agreement across the categories. 
We flag this as an important area to address by future research for performing annotation work with LLMs.

\paragraph{RQ3.} 
This section explores the annotated dataset on \confusion, a category that has high human-model IRR and would be useful for providing instructors feedback. 
Of the total \numComments{} comments, about $16\%$ are annotated with \confusion{} by the model.
The model allows for \textit{easy} identification of areas where students are struggling to understand the material.
This information can be invaluable for teachers looking to refine their instruction (e.g., minimize confusion in their teaching material) and improve the learning experience for their students.
Table~\ref{tab:confusion_comments_diversity} illustrates the diversity in subcategories within \confusion. 
We used the qualitative research approach of grounded theory to discover these categories.
There is a range of comments which ask a conceptual questions (e.g., ``Can anyone explain what professor meant by pivot variables ?'') or express confusion due to a potential mistake in the lecture (e.g., ``i think the explanation of the first question was a little bit wrong it seems.'')
Instructors may use these identified comments to appropriately adapt their lecture content in future course iterations.

\section{Limitations \label{sec:limitations}}

While our work provides a useful starting point for understanding student feedback, there are limitations to our work.
Addressing these limitations will be an important area for future research.

\paragraph{Comments may not reflect real student feedback.} 
The comments in our dataset are from users who have chosen to post publicly on YouTube.
Additionally, the comments may include features specific to this online education setting.
Thus, the comments may reflect real student comments from these courses.

\paragraph{There is a selection bias in lecture sources.}
\datasetname{} includes lectures that may be drawn from the most successful offerings of that course. 
The instructional quality may not be representative of typical instruction.
Thus, inferences drawn about the instruction should be interpreted with caution, as they might not generalize to other lecture settings.

\paragraph{We analyze only English comments.}
We analyze only English comments because the lecture content is given in English and the authors are most comfortable with English.
As a result, our rubric may not capture the types of feedback from non-English students watching lectures taught in English.
In the future, the rubric and analysis should be adapted to account for the multilingual feedback setting.

\paragraph{We annotate a small subsample of the data}
To assess the validity of the automatic labels, we conduct a diagnostic study on a small, randomly selected subset of the dataset, comprising approximately 2\% of the comments. 
Our work aims to establish a preliminary evaluation of the human-model agreement and model annotations, and further validation of the automatic labels is necessary. 
Future work can focus on acquiring such gold-standard annotations to enhance the quality and reliability of the automatic labels.

\section{Future Work}

This work contributes \datasetname{} (a dataset of lecture transcripts and student comments), a rubric for annotating student comments, and an analysis on the annotation quality of LLMs and annotated comments. 

\paragraph{Synthesizing student feedback effectively for instructors.} 
Given the large volume of feedback that instructors receive, it is important to develop methods for summarizing student feedback \citep{hu2022setsum}. 
Equally important is how to present the feedback to teachers such that teachers receive it well and can easily incorporate it into their instruction \citep{yao2005faculty,lindahl2010cruelty}.

\paragraph{Revising the lecture content with respect to student feedback.}

\datasetname{} contains per-lecture user comments. 
This can serve as \textit{language feedback} for revising the lecture content conditioned on student feedback \citep{scheurer2022training}. 

\paragraph{Expanding the human annotations}
Our work relies on two annotators (co-authors) familiar with rubric categories who annotated 280 comments from the total of \numComments{} comments. 
Future work can investigate expanding the human annotations using our rubric, which may be useful for finetuning or evaluations. 
Additionally, the rubric categories focus on the themes that emerge from the comments.
These can act as an initial filter on relevant versus irrelevant comments for instruction feedback.
Future work can consider incorporating categories that play a more specific role for their use case, such as capturing the student's experience in the course \citep{welch-mihalcea-2016-targeted, ganesh-etal-2022-response}.

\paragraph{Improving the model's annotation for qualitative analysis methods} 
Our work shows that the model does not annotate categories well that require more interpretation, even with auxiliary information. 
Future work can explore alternative best practices needed in prompting for these types of categories. 

\section{Conclusion}
Our work contributes \datasetname{}, a large-scale dataset of lecture transcripts and student comments. 
We propose a rubric and different prompting methods for performing automated annotations on \datasetname{}. 
While we find that there is still room for improvement on reliably automating the annotation process, the dataset and rubric provide a foundation for future research to address the challenges of discovering useful feedback from students at scale. 
For qualitative researchers, \datasetname{} offers a unique opportunity to investigate and gain insights into the feedback provided by students in online learning environments. 
The comments cover students' perspectives and opinions related to math lectures. 
Educators can also leverage \datasetname{} as a valuable resource to learn from student comments and refine their teaching materials. 
By analyzing the feedback provided by students, educators can identify strengths and weaknesses in their instruction, discover areas that students find challenging or confusing, and gather valuable insights to enhance their teaching methodologies. 
We hope the dataset and methods building off of this dataset can aid educators in making data-informed decisions to optimize their instructional practices, thereby promoting a more effective learning environment.

\section*{Ethics Statement}

The purpose of this work is to promote better understanding of effective pedagogy and student feedback through the use of NLP techniques.
The \datasetname{} dataset is intended for research purposes only, and is licensed under MIT's Creative Commons License.
The dataset should not be used for commercial purposes, and we ask that users of our dataset respect this restriction.
As stewards of this data, we are committed to protecting the privacy and confidentiality of the individuals who contributed comments to the dataset.
It is important to note that inferences drawn from the dataset may not necessarily reflect the experiences or opinions of real students, and should be interpreted with caution.
The intended use case for this dataset is to further education research and improve teaching and learning outcomes.
Unacceptable use cases include any attempts to identify users or use the data for commercial gain.
We additionally recommend that researchers who do use our dataset take steps to mitigate any risks or harms to individuals that may arise. 

\section*{Acknowledgements}
REW is supported by the National Science Foundation Graduate Research Fellowship. 
We thank Professor Shannon Seidel for feedback during the initial stages of the project. 
We thank Yunsung Kim, Allen Nie, Haley Lepp, and Gabriel Poesia for their feedback on the manuscript. 
We also thank Betsy Rajala, Quinn Waeiss and Professor Michael Bernstein from the Ethics and Society Review Board for guidance on the ethical and safety risks with the data used in our work. 

\bibliography{anthology,custom}

\begin{thebibliography}{50}
\expandafter\ifx\csname natexlab\endcsname\relax\def\natexlab#1{#1}\fi

\bibitem[{Alhija and Fresko(2009)}]{alhija2009student}
Fadia Nasser-Abu Alhija and Barbara Fresko. 2009.
\newblock Student evaluation of instruction: what can be learned from
  students’ written comments?
\newblock \emph{Studies in Educational evaluation}, 35(1):37--44.

\bibitem[{Asselin et~al.(2011)Asselin, Dobson, Meyers, Teixiera, and
  Ham}]{asselin2011learning}
Marlene Asselin, Teresa Dobson, Eric~M Meyers, Cristina Teixiera, and Linda
  Ham. 2011.
\newblock Learning from youtube: an analysis of information literacy in user
  discourse.
\newblock In \emph{Proceedings of the 2011 iConference}, pages 640--642.

\bibitem[{Azab et~al.(2016)Azab, Mihalcea, and Abernethy}]{azab2016analysing}
Mahmoud Azab, Rada Mihalcea, and Jacob Abernethy. 2016.
\newblock Analysing ratemyprofessors evaluations across institutions,
  disciplines, and cultures: The tell-tale signs of a good professor.
\newblock In \emph{Social Informatics: 8th International Conference, SocInfo
  2016, Bellevue, WA, USA, November 11-14, 2016, Proceedings, Part I 8}, pages
  438--453. Springer.

\bibitem[{Baddam et~al.(2019)Baddam, Bingi, and Shuva}]{baddam2019student}
Swathi Baddam, Prasad Bingi, and Syed Shuva. 2019.
\newblock Student evaluation of teaching in business education: Discovering
  student sentiments using text mining techniques.
\newblock \emph{e-Journal of Business Education and Scholarship of Teaching},
  13(3):1--13.

\bibitem[{Bauer and Gaskell(2000)}]{bauer2000qualitative}
Martin~W Bauer and George Gaskell. 2000.
\newblock \emph{Qualitative researching with text, image and sound: A practical
  handbook for social research}.
\newblock Sage.

\bibitem[{Breslow et~al.(2013)Breslow, Pritchard, DeBoer, Stump, Ho, and
  Seaton}]{breslow2013studying}
Lori Breslow, David~E Pritchard, Jennifer DeBoer, Glenda~S Stump, Andrew~D Ho,
  and Daniel~T Seaton. 2013.
\newblock Studying learning in the worldwide classroom research into edx's
  first mooc.
\newblock \emph{Research \& Practice in Assessment}, 8:13--25.

\bibitem[{Brook(2011)}]{brook2011affordances}
Jennifer Brook. 2011.
\newblock The affordances of youtube for language learning and teaching.
\newblock \emph{Hawaii Pacific University TESOL Working Paper Series}, 9(1):2.

\bibitem[{Brown et~al.(2020)Brown, Mann, Ryder, Subbiah, Kaplan, Dhariwal,
  Neelakantan, Shyam, Sastry, Askell et~al.}]{brown2020language}
Tom Brown, Benjamin Mann, Nick Ryder, Melanie Subbiah, Jared~D Kaplan, Prafulla
  Dhariwal, Arvind Neelakantan, Pranav Shyam, Girish Sastry, Amanda Askell,
  et~al. 2020.
\newblock Language models are few-shot learners.
\newblock \emph{Advances in neural information processing systems},
  33:1877--1901.

\bibitem[{Chen and Hoshower(2003)}]{chen2003student}
Yining Chen and Leon~B Hoshower. 2003.
\newblock Student evaluation of teaching effectiveness: An assessment of
  student perception and motivation.
\newblock \emph{Assessment \& evaluation in higher education}, 28(1):71--88.

\bibitem[{Child et~al.(2019)Child, Gray, Radford, and
  Sutskever}]{child2019generating}
Rewon Child, Scott Gray, Alec Radford, and Ilya Sutskever. 2019.
\newblock \href {http://arxiv.org/abs/1904.10509} {Generating long sequences
  with sparse transformers}.

\bibitem[{Chung et~al.(2022)Chung, Hou, Longpre, Zoph, Tay, Fedus, Li, Wang,
  Dehghani, Brahma, Webson, Gu, Dai, Suzgun, Chen, Chowdhery, Castro-Ros,
  Pellat, Robinson, Valter, Narang, Mishra, Yu, Zhao, Huang, Dai, Yu, Petrov,
  Chi, Dean, Devlin, Roberts, Zhou, Le, and Wei}]{chung2022scaling}
Hyung~Won Chung, Le~Hou, Shayne Longpre, Barret Zoph, Yi~Tay, William Fedus,
  Yunxuan Li, Xuezhi Wang, Mostafa Dehghani, Siddhartha Brahma, Albert Webson,
  Shixiang~Shane Gu, Zhuyun Dai, Mirac Suzgun, Xinyun Chen, Aakanksha
  Chowdhery, Alex Castro-Ros, Marie Pellat, Kevin Robinson, Dasha Valter,
  Sharan Narang, Gaurav Mishra, Adams Yu, Vincent Zhao, Yanping Huang, Andrew
  Dai, Hongkun Yu, Slav Petrov, Ed~H. Chi, Jeff Dean, Jacob Devlin, Adam
  Roberts, Denny Zhou, Quoc~V. Le, and Jason Wei. 2022.
\newblock \href {http://arxiv.org/abs/2210.11416} {Scaling
  instruction-finetuned language models}.

\bibitem[{Cohen(1981)}]{cohen1981student}
Peter~A Cohen. 1981.
\newblock Student ratings of instruction and student achievement: A
  meta-analysis of multisection validity studies.
\newblock \emph{Review of educational research}, 51(3):281--309.

\bibitem[{Corbin et~al.(1990)}]{corbin1990basics}
Juliet Corbin et~al. 1990.
\newblock Basics of qualitative research grounded theory procedures and
  techniques.

\bibitem[{Corbin and Strauss(1990)}]{corbin1990grounded}
Juliet~M Corbin and Anselm Strauss. 1990.
\newblock Grounded theory research: Procedures, canons, and evaluative
  criteria.
\newblock \emph{Qualitative sociology}, 13(1):3--21.

\bibitem[{Curran et~al.(2020)Curran, Simmons, Matthews, Fleet, Gustafson,
  Fairbridge, and Xu}]{curran2020youtube}
Vernon Curran, Karla Simmons, Lauren Matthews, Lisa Fleet, Diana~L Gustafson,
  Nicholas~A Fairbridge, and Xiaolin Xu. 2020.
\newblock Youtube as an educational resource in medical education: a scoping
  review.
\newblock \emph{Medical Science Educator}, 30:1775--1782.

\bibitem[{Dubovi and Tabak(2020)}]{dubovi2020empirical}
Ilana Dubovi and Iris Tabak. 2020.
\newblock An empirical analysis of knowledge co-construction in youtube
  comments.
\newblock \emph{Computers \& Education}, 156:103939.

\bibitem[{Erickson et~al.(1985)}]{erickson1985qualitative}
Frederick Erickson et~al. 1985.
\newblock \emph{Qualitative methods in research on teaching}.
\newblock Institute for Research on Teaching.

\bibitem[{Evans and Guymon(1978)}]{evans1978clarity}
Warren~E Evans and Ronald~E Guymon. 1978.
\newblock Clarity of explanation: A powerful indicator of teacher
  effectiveness.

\bibitem[{for Teaching~Project(2011)}]{learning2011measuring}
Learning~Mathematics for Teaching~Project. 2011.
\newblock Measuring the mathematical quality of instruction.
\newblock \emph{Journal of Mathematics Teacher Education}, 14(1):25--47.

\bibitem[{Ganesh et~al.(2022)Ganesh, Scribner, Singh, Goodman, Hertzberg, and
  Kann}]{ganesh-etal-2022-response}
Ananya Ganesh, Hugh Scribner, Jasdeep Singh, Katherine Goodman, Jean Hertzberg,
  and Katharina Kann. 2022.
\newblock \href {https://doi.org/10.18653/v1/2022.bea-1.29} {Response construct
  tagging: {NLP}-aided assessment for engineering education}.
\newblock In \emph{Proceedings of the 17th Workshop on Innovative Use of NLP
  for Building Educational Applications (BEA 2022)}, pages 250--261, Seattle,
  Washington. Association for Computational Linguistics.

\bibitem[{Gilardi et~al.(2023)Gilardi, Alizadeh, and
  Kubli}]{gilardi2023chatgpt}
Fabrizio Gilardi, Meysam Alizadeh, and Maël Kubli. 2023.
\newblock \href {http://arxiv.org/abs/2303.15056} {Chatgpt outperforms
  crowd-workers for text-annotation tasks}.

\bibitem[{Gottipati et~al.(2018)Gottipati, Shankararaman, and
  Lin}]{gottipati2018text}
Swapna Gottipati, Venky Shankararaman, and Jeff~Rongsheng Lin. 2018.
\newblock Text analytics approach to extract course improvement suggestions
  from students’ feedback.
\newblock \emph{Research and Practice in Technology Enhanced Learning},
  13:1--19.

\bibitem[{Gravestock and Gregor-Greenleaf(2008)}]{gravestock2008student}
Pamela Gravestock and Emily Gregor-Greenleaf. 2008.
\newblock \emph{Student course evaluations: Research, models and trends}.
\newblock Higher Education Quality Council of Ontario Toronto.

\bibitem[{Greenwald and Gillmore(1997)}]{greenwald1997grading}
Anthony~G Greenwald and Gerald~M Gillmore. 1997.
\newblock Grading leniency is a removable contaminant of student ratings.
\newblock \emph{American psychologist}, 52(11):1209.

\bibitem[{Hammonds et~al.(2017)Hammonds, Mariano, Ammons, and
  Chambers}]{hammonds2017student}
Frank Hammonds, Gina~J Mariano, Gracie Ammons, and Sheridan Chambers. 2017.
\newblock Student evaluations of teaching: improving teaching quality in higher
  education.
\newblock \emph{Perspectives: Policy and Practice in Higher Education},
  21(1):26--33.

\bibitem[{Harrison et~al.(2019)Harrison, Nguyen, Seidel, Escobedo, Hartman,
  Lam, Liang, Martens, Acker, Akana et~al.}]{harrison2019investigating}
Colin~D Harrison, Tiffy~A Nguyen, Shannon~B Seidel, Alycia~M Escobedo, Courtney
  Hartman, Katie Lam, Kristen~S Liang, Miranda Martens, Gigi~N Acker, Susan~F
  Akana, et~al. 2019.
\newblock Investigating instructor talk in novel contexts: Widespread use,
  unexpected categories, and an emergent sampling strategy.
\newblock \emph{CBE—Life Sciences Education}, 18(3):ar47.

\bibitem[{Hativa(1998)}]{hativa1998lack}
Nira Hativa. 1998.
\newblock Lack of clarity in university teaching: A case study.
\newblock \emph{Higher Education}, pages 353--381.

\bibitem[{He et~al.(2023)He, Lin, Gong, Jin, Zhang, Lin, Jiao, Yiu, Duan, and
  Chen}]{he2023annollm}
Xingwei He, Zhenghao Lin, Yeyun Gong, A-Long Jin, Hang Zhang, Chen Lin, Jian
  Jiao, Siu~Ming Yiu, Nan Duan, and Weizhu Chen. 2023.
\newblock \href {http://arxiv.org/abs/2303.16854} {Annollm: Making large
  language models to be better crowdsourced annotators}.

\bibitem[{Hu et~al.(2022)Hu, Zhang, Sathy, Panter, and Bansal}]{hu2022setsum}
Yinuo Hu, Shiyue Zhang, Viji Sathy, AT~Panter, and Mohit Bansal. 2022.
\newblock Setsum: Summarization and visualization of student evaluations of
  teaching.
\newblock \emph{arXiv preprint arXiv:2207.03640}.

\bibitem[{Hujala et~al.(2020)Hujala, Knutas, Hynninen, and
  Arminen}]{hujala2020improving}
Maija Hujala, Antti Knutas, Timo Hynninen, and Heli Arminen. 2020.
\newblock Improving the quality of teaching by utilising written student
  feedback: A streamlined process.
\newblock \emph{Computers \& education}, 157:103965.

\bibitem[{Hynninen et~al.(2019)Hynninen, Knutas, Hujala, and
  Arminen}]{hynninen2019distinguishing}
Timo Hynninen, Antti Knutas, Maija Hujala, and Heli Arminen. 2019.
\newblock Distinguishing the themes emerging from masses of open student
  feedback.
\newblock In \emph{2019 42nd International Convention on Information and
  Communication Technology, Electronics and Microelectronics (MIPRO)}, pages
  557--561. IEEE.

\bibitem[{Kim and Piech(2023)}]{kim2023high}
Yunsung Kim and Chris Piech. 2023.
\newblock High-resolution course feedback: Timely feedback for course
  instructors.

\bibitem[{Kuzman et~al.(2023)Kuzman, Mozetic, and
  Ljube{\v{s}}ic}]{kuzman2023chatgpt}
Taja Kuzman, Igor Mozetic, and Nikola Ljube{\v{s}}ic. 2023.
\newblock Chatgpt: Beginning of an end of manual linguistic data annotation?
  use case of automatic genre identification.
\newblock \emph{arXiv e-prints}, pages arXiv--2303.

\bibitem[{Lee et~al.(2017)Lee, Osop, Goh, and Kelni}]{lee2017making}
Chei~Sian Lee, Hamzah Osop, Dion Hoe-Lian Goh, and Gani Kelni. 2017.
\newblock Making sense of comments on youtube educational videos: A
  self-directed learning perspective.
\newblock \emph{Online Information Review}, 41(5):611--625.

\bibitem[{Lee et~al.(2022)Lee, Ahuja, Liang, Natu, and
  Morency}]{lee2022multimodal}
Dong~Won Lee, Chaitanya Ahuja, Paul~Pu Liang, Sanika Natu, and Louis-Philippe
  Morency. 2022.
\newblock Multimodal lecture presentations dataset: Understanding multimodality
  in educational slides.
\newblock \emph{arXiv preprint arXiv:2208.08080}.

\bibitem[{Lindahl and Unger(2010)}]{lindahl2010cruelty}
Mary~W Lindahl and Michael~L Unger. 2010.
\newblock Cruelty in student teaching evaluations.
\newblock \emph{College Teaching}, 58(3):71--76.

\bibitem[{Marsh and Roche(1997)}]{marsh1997making}
Herbert~W Marsh and Lawrence~A Roche. 1997.
\newblock Making students' evaluations of teaching effectiveness effective: The
  critical issues of validity, bias, and utility.
\newblock \emph{American psychologist}, 52(11):1187.

\bibitem[{OCW(2020)}]{mit_2020}
MIT OCW. 2020.
\newblock 2020 ocw impact report.
\newblock \url{https://ocw.mit.edu/ocw-www/2020-19_ocw_impact_report.pdf}.

\bibitem[{OCW(2023)}]{mit_ocw}
MIT OCW. 2023.
\newblock Massachusetts institute of technology: Mit opencouseware.
\newblock \url{https://ocw.mit.edu/}.

\bibitem[{OpenAI(2023)}]{openai_chatgpt_2023}
OpenAI. 2023.
\newblock Introducing chatgpt and whisper apis.
\newblock \url{https://openai.com/blog/introducing-chatgpt-and-whisper-apis}.

\bibitem[{Ouyang et~al.(2022)Ouyang, Wu, Jiang, Almeida, Wainwright, Mishkin,
  Zhang, Agarwal, Slama, Ray, Schulman, Hilton, Kelton, Miller, Simens, Askell,
  Welinder, Christiano, Leike, and Lowe}]{ouyang2022training}
Long Ouyang, Jeff Wu, Xu~Jiang, Diogo Almeida, Carroll~L. Wainwright, Pamela
  Mishkin, Chong Zhang, Sandhini Agarwal, Katarina Slama, Alex Ray, John
  Schulman, Jacob Hilton, Fraser Kelton, Luke Miller, Maddie Simens, Amanda
  Askell, Peter Welinder, Paul Christiano, Jan Leike, and Ryan Lowe. 2022.
\newblock \href {http://arxiv.org/abs/2203.02155} {Training language models to
  follow instructions with human feedback}.

\bibitem[{O’Connor and Joffe(2020)}]{o2020intercoder}
Cliodhna O’Connor and Helene Joffe. 2020.
\newblock Intercoder reliability in qualitative research: debates and practical
  guidelines.
\newblock \emph{International journal of qualitative methods},
  19:1609406919899220.

\bibitem[{Pianta et~al.(2008)Pianta, La~Paro, and Hamre}]{pianta2008classroom}
Robert~C Pianta, Karen~M La~Paro, and Bridget~K Hamre. 2008.
\newblock \emph{Classroom Assessment Scoring System™: Manual K-3.}
\newblock Paul H Brookes Publishing.

\bibitem[{Radford et~al.(2022)Radford, Kim, Xu, Brockman, McLeavey, and
  Sutskever}]{radford2022robust}
Alec Radford, Jong~Wook Kim, Tao Xu, Greg Brockman, Christine McLeavey, and
  Ilya Sutskever. 2022.
\newblock \href {http://arxiv.org/abs/2212.04356} {Robust speech recognition
  via large-scale weak supervision}.

\bibitem[{Scheurer et~al.(2022)Scheurer, Campos, Chan, Chen, Cho, and
  Perez}]{scheurer2022training}
Jérémy Scheurer, Jon~Ander Campos, Jun~Shern Chan, Angelica Chen, Kyunghyun
  Cho, and Ethan Perez. 2022.
\newblock \href {http://arxiv.org/abs/2204.14146} {Training language models
  with language feedback}.

\bibitem[{Seidel et~al.(2015)Seidel, Reggi, Schinske, Burrus, and
  Tanner}]{seidel2015beyond}
Shannon~B Seidel, Amanda~L Reggi, Jeffrey~N Schinske, Laura~W Burrus, and
  Kimberly~D Tanner. 2015.
\newblock Beyond the biology: A systematic investigation of noncontent
  instructor talk in an introductory biology course.
\newblock \emph{CBE—Life Sciences Education}, 14(4):ar43.

\bibitem[{Welch and Mihalcea(2016)}]{welch-mihalcea-2016-targeted}
Charles Welch and Rada Mihalcea. 2016.
\newblock \href {https://aclanthology.org/C16-1233} {Targeted sentiment to
  understand student comments}.
\newblock In \emph{Proceedings of {COLING} 2016, the 26th International
  Conference on Computational Linguistics: Technical Papers}, pages 2471--2481,
  Osaka, Japan. The COLING 2016 Organizing Committee.

\bibitem[{Yao and Grady(2005)}]{yao2005faculty}
Yuankun Yao and Marilyn~L Grady. 2005.
\newblock How do faculty make formative use of student evaluation feedback?: A
  multiple case study.
\newblock \emph{Journal of Personnel Evaluation in Education}, 18:107--126.

\bibitem[{Zabaleta(2007)}]{zabaleta2007use}
Francisco Zabaleta. 2007.
\newblock The use and misuse of student evaluations of teaching.
\newblock \emph{Teaching in higher education}, 12(1):55--76.

\bibitem[{Ziems et~al.(2023)Ziems, Held, Shaikh, Chen, Zhang, and
  Yang}]{ziems2023css}
Caleb Ziems, William Held, Omar Shaikh, Jiaao Chen, Zhehao Zhang, and Diyi
  Yang. 2023.
\newblock Can large language models transform computational social science?
\newblock \emph{arXiv submission 4840038}.

\end{thebibliography}
\bibliographystyle{acl_natbib}

\appendix

\section{\datasetname{} details \label{sec:app_dataset}}

The playlists used in \datasetname{} are:
\begin{itemize}
    \item 6.041 Probabilistic Systems Analysis and Applied Probability
    \item 18.01 Single Variable Calculus
    \item 18.02 Multivariable Calculus
    \item 18.404J Theory of Computation
    \item 18.06 Linear Algebra
    \item 18.065 Matrix Methods in Data Analysis, Signal Processing, and Machine Learning
    \item 18.100A Real Analysis
    \item 18.102 Introduction to Functional Analysis
    \item 18.217 Graph Theory and Additive Combinatorics
    \item 18.650 Statistics for Applications
\end{itemize}

Table~\ref{tab:number_of_comments} shows each playlist's number of comments.
These comments are collected through Google's YouTube API as detailed in Section~\ref{sec:dataset}.

\begin{table*}
\centering
\small
\begin{tabular}{l|c}
\toprule
\textbf{Playlist name} & \textbf{Number of comments}\\
\midrule
6.041 Probabilistic Systems Analysis and Applied Probability & 1,031 \\
\midrule
18.01 Single Variable Calculus & 3,293 \\
\midrule
18.02 Multivariable Calculus & 2,642 \\
\midrule
18.404J Theory of Computation & 202 \\
\midrule
18.06 Linear Algebra & 6,021 \\
\midrule
18.065 Matrix Methods in Data Analysis, Signal Processing, and Machine Learning & 1,448 \\
\midrule
18.100A Real Analysis & 244 \\
\midrule
18.102 Introduction to Functional Analysis & 129 \\
\midrule
18.217 Graph Theory and Additive Combinatorics & 78 \\
\midrule
18.650 Statistics for Applications & 696 \\
\bottomrule
\end{tabular}
\caption{Number of comments from each playlist in our dataset.}
\label{tab:number_of_comments}
\end{table*}

\section{Human annotation interface \label{app:human_annotation}}

Figure~\ref{fig:annotation_interface} shows an example of what the annotation interface looks like. 
Each comment had to be labeled with at least one category in order to proceed to the next comment. 
The human annotators annotated the same comments. 
A total of 290 comments are manually annotated and are used to perform the annotation analysis in Section~\ref{sec:analysis}.

\section{Distribution of annotations \label{sec:category_distribution}}
\begin{figure*}[h]
    \newcommand{\ratio}{0.30}
    \centering
    \begin{subfigure}{\linewidth}
        \includegraphics[width=\linewidth]{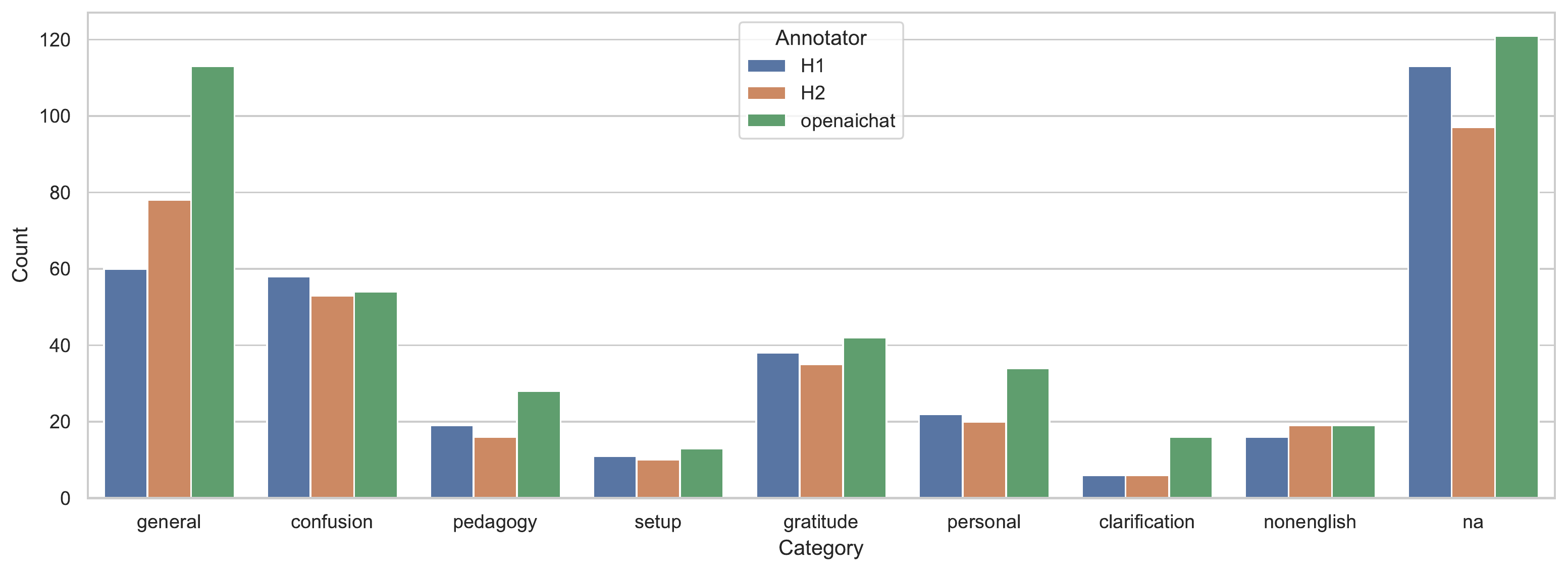}
        \caption{Zero-shot}
    \end{subfigure}
    \begin{subfigure}{\linewidth}
        \includegraphics[width=\linewidth]{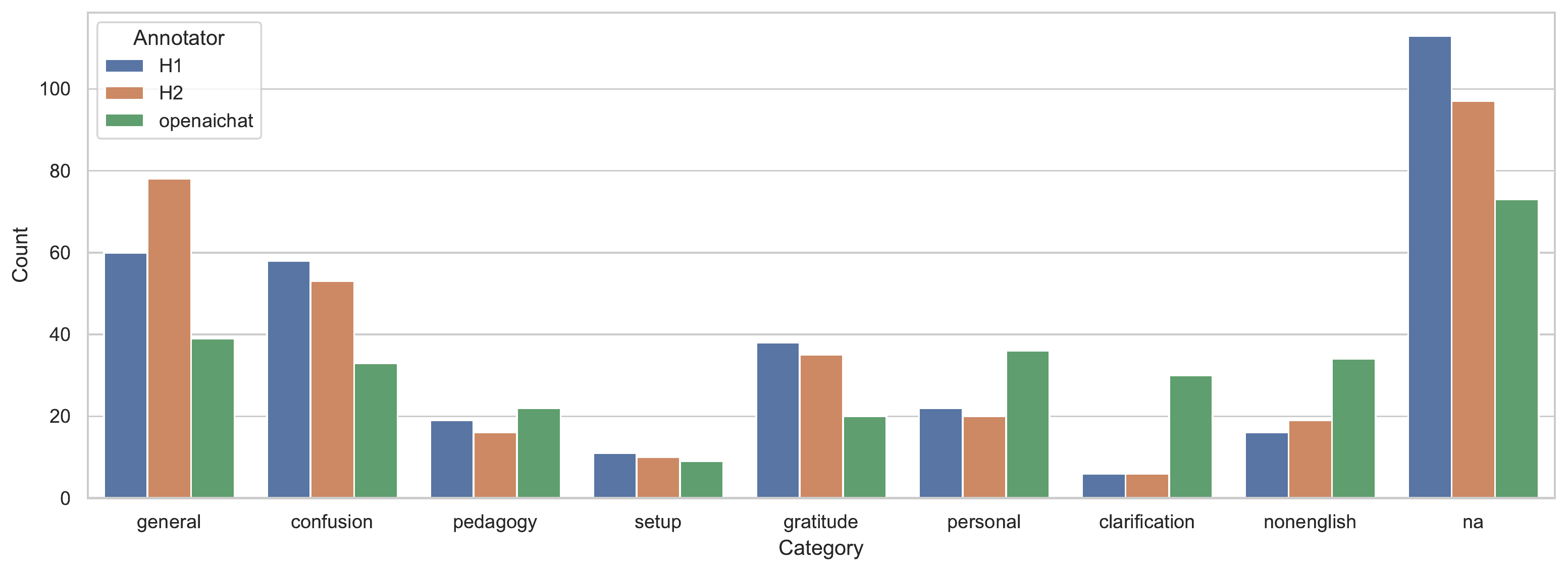}
        \caption{K-shot}
    \end{subfigure}
    \begin{subfigure}{\linewidth}
        \includegraphics[width=\linewidth]{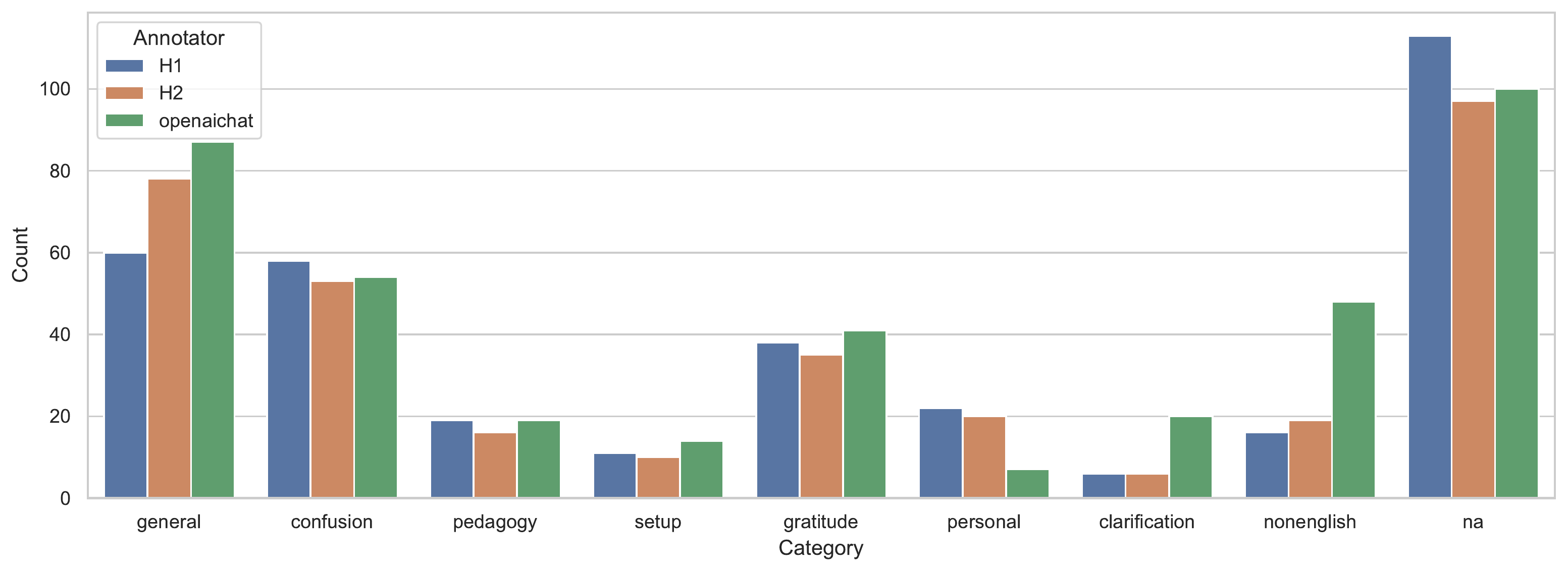}
        \caption{K-shot with reasoning}
    \end{subfigure}
    \caption{
    The distribution of categories annotated by the two humans H1 and H2 as well as (a) the 0-shot prompted model, (b) the k-shot prompted model, and (c) the k-shot with reasoning prompted model.
    \label{fig:distribution_annotation}}
\end{figure*}

Figure~\ref{fig:distribution_annotation} shows the distribution of comment categories from the two human annotators and model.

\begin{figure*}
    \centering 
    \includegraphics[width=\linewidth]{images/category_distribution_bin_zeroshot.pdf}
    \includegraphics[width=\linewidth]{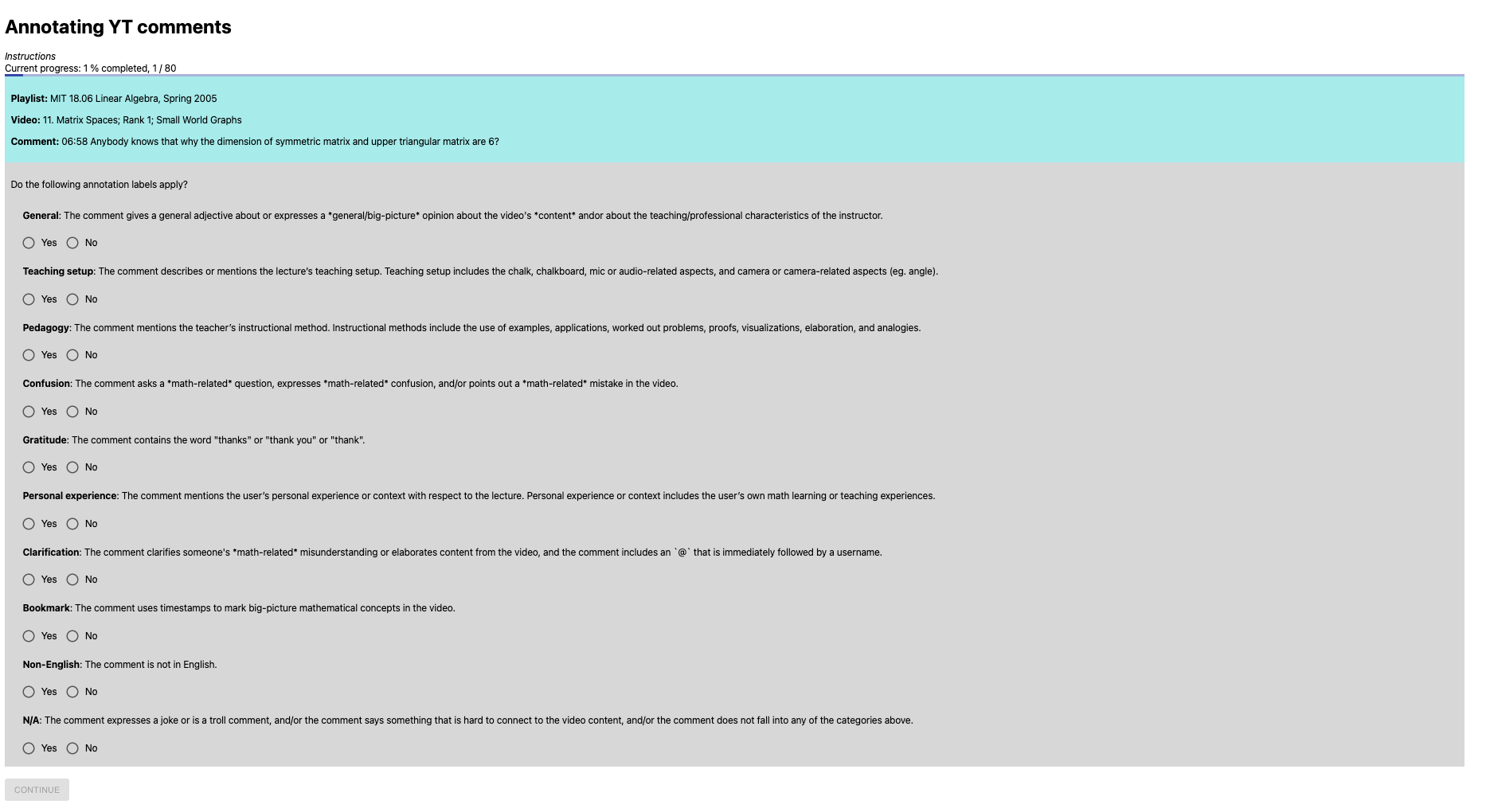}
    \caption{Human annotation interface for labelling YouTube comments
    \label{fig:annotation_interface}}
\end{figure*}

\begin{figure*}
    \centering 
    \includegraphics[width=\linewidth]{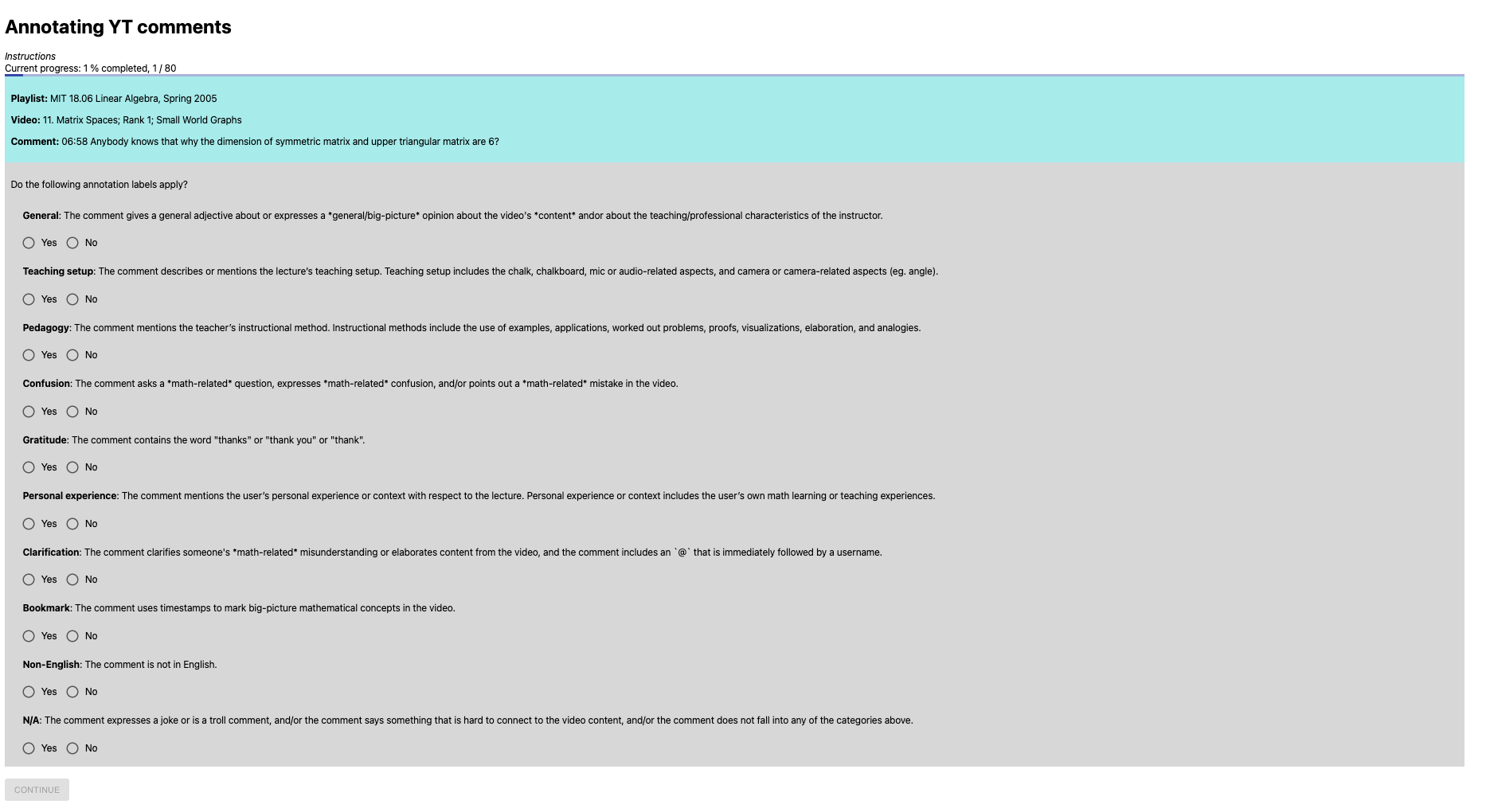}
    \caption{Human annotation interface for labelling YouTube comments
    \label{fig:annotation_interface}}
\end{figure*}

\section{Scaling annotation \label{app:scaling_annotation}}

This section documents prior attempts at scaling annotation with LLMs. 
We believe this is highly instructive for researchers applying LLMs to other domains to facilitate their qualitative analysis. 

\subsection{Multi-class classification with entire rubric \label{app:multiclass_entire_rubric}} 
\paragraph{Setup}
We first attempted to perform multi-class classification over all 9 categories on each comment, i.e., which one category best applies to this comment?
Note this is different from the multi-label classification scheme that our work performs in the main text.
This method similarly provides the context of the comment and the comment itself, then provide instructions on the labelling task. 
The context of the comment includes a mention to MIT OCW, the playlist name, and the video name. 
The instructions include the entire annotation rubric: a list of the all category names and descriptions.
It ends by instructing the model to respond with the category that best applies to the comment. 
An example of such a multi-class classification prompt is shown in Figure \ref{fig:prior_attempt_multiclass}. 

\paragraph{Results}
First, the human-model agreement scores were generally moderate ($\sim 0.50$). 
We found that the \textbf{model did not follow constraints we had set for some categories.}
One constraint is on \gratitude: label comments as \gratitude{} if and only if they contain ``thanks'' or ``thank''. 
We found that the model would still label comments that alluded to being grateful but did not follow this constraint as \gratitude{}. 
Attempts at tuning the prompt did not result in higher IRR.

Additionally, \textbf{a comment may belong to multiple categories in our rubric, making it challenging to make the model to just the one best category for the comment}. 
The human annotators did have rules for resolving category conflicts, i.e., when a comment belongs to more than one category, which of the categories to assign.
These rules were also included in the prompt, however this did not improve the model's annotations much.

\begin{figure*}[t]
    \centering 
    \begin{tcolorbox}[
    modelresponse,
    title={\textbf{Multi-class classification with entire rubric}},
    ]
    \small
Consider a YouTube comment from the math MIT OCW video below:\\
Playlist name: \{playlistName\}\\
Video name: \{videoName\}\\

Your task is to annotate the comment with one of the labels defined below. If the comment fits into multiple labels, please choose the label that best fits the spirit of the comment.\\
- `thanks': The student explicitly expresses gratitude. Comments must include "thanks" or "thank" to be labeled as `thanks'.\\
- `general': The student makes a general comment about the video's *content* or a comment about the *teaching characteristics* of the instructor. If the comment is not related to the content or teaching characteristics (e.g., the comment is about the instructor's appearance or accent), then the comment should be labeled as `na'.\\
- `style': The student comments on *how* the teacher teaches the content. This includes comments on the use of examples, applications, or step-by-step explanations.\\
- `personal\_experience': The student shares a personal experience related to the content, such as their previous attempts at learning the content.\\
- `question': The student expresses *math-related* confusion. Comments must include a question and math-related confusion to be labeled as `question'.\\ 
- `assist': The student clarifies someone's *math-related* misunderstanding or elaborates the content. Comments must include `@`user and math-related content to be labeled as `assist'. \\
- `bookmark': The student uses timestamps to mark *content-related* features of the video, such as the content outline or the start of a topic. Comments must include a timestamp and math-related content to be labeled as `bookmark'. Otherwise (e.g., if a student uses a timestamp to mark a joke), the comment should be labeled as `na'.\\
- `non\_english': The student's comment is not in English. \\
- `na': The student does not express any of the above types of comments. Instead, for example, the student jokes, says something insulting about the content or instructor, or says something unrelated to the math content.\\

Comment: \{comment\}\\
Label:
    \end{tcolorbox}
    \caption{Example of a multi-class classification prompt with entire rubric.
    \label{fig:prior_attempt_multiclass}}
\end{figure*}

\subsection{Staged divide-and-conquer multi-class classification}
\paragraph{Setup}
To more directly help LLMs choose the one label that best applies to a comment, we give the LLM a series of simpler classification subtasks, i.e. a staged classification scheme. 
In the first stage, we ask the model to annotate categories that are more easily resolvable, e.g., \gratitude{}, which only looks for the words ``thanks'' or ``thank'' in comments. 
We provide the model the option to also annotate with ``none'' if none of the of categories in the stage apply.
These ``none'' comments then transition to the next stage of labelling.
The following stages ask the LLM to classify between a set of categories that are often mistaken for each other, and continue to pass comments that have not gotten a label to the following stage(s). 
One example of categories that the LLM frequently mistakes for each other are \general{} and \pedagogy{}: Although both categories can include comments about the teacher, \pedagogy{} should only include comments that explicitly talk about the teacher's instructional method, while \general{} should include comments that expresses more general opinions about the teacher. 
Therefore, we tried grouping \general{} and \pedagogy{} together in one stage in hopes of helping the model more clearly see the difference between the two categories.

\paragraph{Results} We found that compared to the previous classification scheme (Section~\ref{app:multiclass_entire_rubric}), \textbf{asking the LLM to classify between a smaller set of categories that are often mistaken for each other does reduce the number of errors that the LLM makes between those categories}. For example, from frequently mistaking \pedagogy{} as \general{} comments when prompted with the entire rubric (i.e., all other categories), isolating the classification between just \general{} and \pedagogy{} to one stage helped the LLM more accurately decide between labelling a comment as \general{}, \pedagogy{}, or none of the two. However, the human-model agreement scores were still at most moderate ($\sim 0.60$).

Our attempts to perform multi-class classification (with the entire rubric and with the staged divide-and-conquer method) led us to hypothesize that in our setting, the task of choosing only one label that best applies to a comment is too difficult for LLMs to perform reliably.
This is especially the case when comments require using one of our category conflict rules: 
ChatGPT did not seem to handle category conflicts as we had instructed in the prompt.

\subsection{Binary classification per category}
This is the final classification scheme we tried and use in our main work. 

\paragraph{Motivation}
Given the difficulty in performing multi-class classification and given that many comments do fit into multiple categories, we decided to implement binary classification per category.
This involves prompting for each category on whether each label applies to a comment.
This annotation scheme also allows each comment to be labelled with more than one category.

Additionally, in previous attempts, we found that the \textbf{label names in the prompt affects the LLM’s response}.
For example, renaming the \gratitude{} category as ``thanks'' increases the human-model agreement score because the model would otherwise mark every comment that alludes to a grateful sentiment as \gratitude. 
Our binary classification approach eliminates the need for label names in prompts and reduces the priors that LLMs may have on certain label names.

\paragraph{Setup}
As our final prompting strategy, each comment is annotated as a binary classification task per category, i.e., does this category apply to this comment? Note that we have moved from multi-class classification (giving each comment just a single label) to multi-label classification (allowing each comment to get multiple labels). Details on this final prompting strategy can be found in Section~\ref{prompting_methods}. However, note that a category's description in the final version of our prompts may be slightly different from the human description for the category, since after multiple prompt engineering attempts, we found those new descriptions to be better at helping the LLM to detect comments in that category. For example, to help the LLM label short adjective comments like ``Marvelous!!!'' as \general{}, we had to add the underlined part to the \general{} category's description in our \general{} prompt: ``The comment \underline{expresses a general sentiment/adjective about or} expresses a general/big-picture opinion about the video's content and/or about the teaching/professional characteristics of the instructor.''

\paragraph{Results} Results for this binary classification per category method can be found in Section~\ref{sec:analysis}.

\section{Prompts \label{app:all_prompts}} 

This section details the prompts used for the 0-shot, k-shot and k-shot reasoning prompting strategies. 
We use these prompts to get the model annotations used in Section~\ref{sec:analysis}.

\subsection{0-shot prompts}

\begin{itemize}
    \item Figure~\ref{fig:zeroshot_general_prompt} is the 0-shot prompt for \general.
    \item Figure~\ref{fig:zeroshot_confusion_prompt} is the 0-shot prompt for \confusion.
    \item Figure~\ref{fig:zeroshot_pedagogy_prompt} is the 0-shot prompt for \pedagogy.
    \item Figure~\ref{fig:zeroshot_setup_prompt} is the 0-shot prompt for \setup.
    \item Figure~\ref{fig:zeroshot_personal_prompt} is the 0-shot prompt for \personal.
    \item Figure~\ref{fig:zeroshot_clarification_prompt} is the 0-shot prompt for \clarification.
    \item Figure~\ref{fig:zeroshot_gratitude_prompt} is the 0-shot prompt for \gratitude.
    \item Figure~\ref{fig:zeroshot_nonenglish_prompt} is the 0-shot prompt for \nonenglish.
    \item Figure~\ref{fig:zeroshot_na_prompt} is the 0-shot prompt for \na.
\end{itemize}

\begin{figure*}[t]
    \centering 
    \begin{tcolorbox}[
    zeroshot,
    title={\small \textbf{Zero-shot prompting for \general{} category}},
    ]
    \small 
    Consider a YouTube comment from the math MIT OCW video below:\\
Playlist name: \{playlistName\}\\
Video name: \{videoName\}\\
Comment: \{comment\}\\

If the statement below is true, please respond "true"; otherwise, please respond "false":\\
The comment expresses a general sentiment/adjective about or expresses a *general/big-picture* opinion about the video's *content* and/or about the teaching/professional characteristics of the *instructor*.
    \end{tcolorbox}
    \vspace{-1em}
    \caption{
    The zero-shot prompt for the \general{} category.
    \label{fig:zeroshot_general_prompt}
    }
\end{figure*}

\begin{figure*}[t]
    \centering 
    \begin{tcolorbox}[
    zeroshot,
    title={\small \textbf{Zero-shot prompting for \confusion{} category}},
    ]
    \small 
    Consider a YouTube comment from the math MIT OCW video below: \\
Playlist name: \{playlistName\}\\
Video name: \{videoName\}\\
Comment: \{comment\}\\

If the statement below is true, please respond "true"; otherwise, please respond "false":\\
The comment asks a specific mathematical question and/or points out a mathematical mistake in the video.
    \end{tcolorbox}
    \vspace{-1em}
    \caption{
    The zero-shot prompt for the \confusion{} category.
    \label{fig:zeroshot_confusion_prompt}
    }
\end{figure*}

\begin{figure*}[t]
    \centering 
    \begin{tcolorbox}[
    zeroshot,
    title={\small \textbf{Zero-shot prompting for \pedagogy{} category}},
    ]
    \small 
    Consider a YouTube comment from the math MIT OCW video below:\\
    
Playlist name: \{playlistName\}\\
Video name: \{videoName\}\\
Comment: \{comment\} \\

If the statement below is true, please respond "true"; otherwise, please respond "false": 

The comment mentions the teacher’s instructional method, which includes but is not limited to the use of examples, applications, worked out problems, proofs, visualizations, elaboration, and analogies.

    \end{tcolorbox}
    \vspace{-1em}
    \caption{
    The zero-shot prompt for the \pedagogy{} category.
    \label{fig:zeroshot_pedagogy_prompt}
    }
\end{figure*}
\begin{figure*}[t]
    \centering 
    \begin{tcolorbox}[
    zeroshot,
    title={\small \textbf{Zero-shot prompting for \setup{} category}},
    ]
    \small 
    Consider a YouTube comment from the math MIT OCW video below:\\
Playlist name: \{playlistName\}\\
Video name: \{videoName\}\\
Comment: \{comment\}\\

If the statement below is true, please respond "true"; otherwise, please respond "false":\\
The comment mentions the lecture’s physical teaching setup, which includes but is not limited to the chalk, board, microphone or audio-related aspects, and camera-related aspects (e.g., angle).
    \end{tcolorbox}
    \vspace{-1em}
    \caption{
    The zero-shot prompt for the \setup{} category.
    \label{fig:zeroshot_setup_prompt}
    }
\end{figure*}
\begin{figure*}[t]
    \centering 
    \begin{tcolorbox}[
    zeroshot,
    title={\small \textbf{Zero-shot prompting for \personal{} category}},
    ]
    \small 
    Consider a YouTube comment from the math MIT OCW video below:\\
Playlist name: \{playlistName\}\\
Video name: \{videoName\}\\
Comment: \{comment\}\\

If the statement below is true, please respond "true"; otherwise, please respond "false":\\
The comment mentions the user’s personal experience learning or teaching math on their own outside of watching this lecture/series.
    \end{tcolorbox}
    \vspace{-1em}
    \caption{
    The zero-shot prompt for the \personal{} category.
    \label{fig:zeroshot_personal_prompt}
    }
\end{figure*}
\begin{figure*}[t]
    \centering 
    \begin{tcolorbox}[
    zeroshot,
    title={\small \textbf{Zero-shot prompting for \clarification{} category}},
    ]
    \small 
    Consider a YouTube comment from the math MIT OCW video below: \\ 
Playlist name: \{playlistName\}\\
Video name: \{videoName\}\\
Comment: \{comment\}\\

If the statement below is true, please respond "true"; otherwise, please respond "false": \\ 
The comment clarifies someone's *math-related* misunderstanding or elaborates content from the video, and the comment includes an `@' that is immediately followed by a username.
    \end{tcolorbox}
    \vspace{-1em}
    \caption{
    The zero-shot prompt for the \clarification{} category.
    \label{fig:zeroshot_clarification_prompt}
    }
\end{figure*}

\begin{figure*}[t]
    \centering 
    \begin{tcolorbox}[
    zeroshot,
    title={\small \textbf{Zero-shot prompting for \gratitude{} category}},
    ]
    \small 
    Consider a YouTube comment from the math MIT OCW video below: \\
Playlist name: \{playlistName\}\\
Video name: \{videoName\}\\
Comment: \{comment\}\\

If the statement below is true, please respond "true"; otherwise, please respond "false": \\
The comment contains the word "thanks" or "thank". \\
    \end{tcolorbox}
    \vspace{-1em}
    \caption{
    The zero-shot prompt for the \gratitude{} category.
    \label{fig:zeroshot_gratitude_prompt}
    }
\end{figure*}
\begin{figure*}[t]
    \centering 
    \begin{tcolorbox}[
    zeroshot,
    title={\small \textbf{Zero-shot prompting for \nonenglish{} category}},
    ]
    \small 
    Consider a YouTube comment from the math MIT OCW video below:\\
Playlist name: \{playlistName\}\\
Video name: \{videoName\}\\
Comment: \{comment\}\\

If the statement below is true, please respond "true"; otherwise, please respond "false":\\
The comment is in English.
    \end{tcolorbox}
    \vspace{-1em}
    \caption{
    The zero-shot prompt for the \nonenglish{} category. The final label on this is flipped.
    \label{fig:zeroshot_nonenglish_prompt}
    }
\end{figure*}
\begin{figure*}[t]
    \centering 
    \begin{tcolorbox}[
    zeroshot,
    title={\small \textbf{Zero-shot prompting for \na{} category}},
    ]
    \small 
    Consider a YouTube comment from the math MIT OCW video below:\\
Playlist name: \{playlistName\}\\
Video name: \{videoName\}\\
Comment: \{comment\}\\

If the statement below is true, please respond "true"; otherwise, please respond "false":\\
The comment expresses a joke or is a troll comment.\\
    \end{tcolorbox}
    \vspace{-1em}
    \caption{
    The zero-shot prompt for the \na{} category.
    \label{fig:zeroshot_na_prompt}
    }
\end{figure*}

\subsection{k-shot prompts}

\begin{itemize}
    \item Figure~\ref{fig:kshot_general_prompt} is the k-shot prompt for \general.
    \item Figure~\ref{fig:kshot_confusion_prompt} is the k-shot prompt for \confusion.
    \item Figure~\ref{fig:kshot_pedagogy_prompt} is the k-shot prompt for \pedagogy.
    \item Figure~\ref{fig:kshot_setup_prompt} is the k-shot prompt for \setup.
    \item Figure~\ref{fig:kshot_personal_prompt} is the k-shot prompt for \personal.
    \item Figure~\ref{fig:kshot_clarification_prompt} is the k-shot prompt for \clarification.
    \item Figure~\ref{fig:kshot_gratitude_prompt} is the k-shot prompt for \gratitude.
    \item Figure~\ref{fig:kshot_nonenglish_prompt} is the k-shot prompt for \nonenglish.
    \item Figure~\ref{fig:kshot_na_prompt} is the k-shot prompt for \na.
\end{itemize}

\begin{figure*}[t]
    \centering 
    \begin{tcolorbox}[
    kshot,
    title={\small \textbf{K-shot  prompting for \general{} category}},
    ]
    \small 
    Given a user comment on YouTube from a math MIT OCW video, your task is to label whether the comment expresses a general sentiment/adjective about or expresses a *general/big-picture* opinion about the video's *content* and/or about the teaching/professional characteristics of the *instructor*. If it is true, then label "true"; otherwise, label "false". \\

Consider a YouTube comment from the math MIT OCW video below:\\
Playlist name: MIT 18.06 Linear Algebra, Spring 2005\\
Video name: 34. Final Course Review\\
Comment: Absolutely well done and definitely keep it up!!! :thumbs\_up::thumbs\_up:\\:thumbs\_up::thumbs\_up::thumbs\_up::thumbs\_up::thumbs\_up::thumbs\_up:\\:thumbs\_up::thumbs\_up::thumbs\_up::thumbs\_up::thumbs\_up::thumbs\_up:\\
Task: Does the comment express a general opinion about the video's content and/or about the teaching/professional characteristics of the instructor?\\
Label: true\\

Consider a YouTube comment from the math MIT OCW video below:\\
Playlist name: MIT 18.02 Multivariable Calculus, Fall 2007\\
Video name: Lec 3: Matrices; inverse matrices | MIT 18.02 Multivariable Calculus, Fall 2007 \\
Comment: Ideally, do you learn multivariable calculus first or linear algebra? A lot of stuff here seems to be based on 18.06.\\
Task: Does the comment express a general opinion about the video's content and/or about the teaching/professional characteristics of the instructor?\\
Label: false\\

Consider a YouTube comment from the math MIT OCW video below:\\
Playlist name: MIT 18.02 Multivariable Calculus, Fall 2007\\
Video name: Lec 16: Double integrals | MIT 18.02 Multivariable Calculus, Fall 2007\\
Comment: This video is very helpful, i appreciate the help.\\
Task: Does the comment express a general opinion about the video's content and/or about the teaching/professional characteristics of the instructor?\\
Label: true\\

Consider a YouTube comment from the math MIT OCW video below:\\
Playlist name: \{playlistName\}\\
Video name: \{videoName\}\\
Comment: \{comment\}\\
Task: Does the comment express a general opinion about the video's content and/or about the teaching/professional characteristics of the instructor?\\
Label:
    \end{tcolorbox}
    \vspace{-1em}
    \caption{
    The k-shot prompt for the \general{} category.
    \label{fig:kshot_general_prompt}
    }
\end{figure*}
\begin{figure*}[t]
    \centering 
    \begin{tcolorbox}[
    kshot,
    title={\small \textbf{K-shot  prompting for \confusion{} category}},
    ]
    \small 
    Given a user comment on YouTube from a math MIT OCW video, your task is to label whether the comment asks a specific mathematical question and/or points out a mathematical mistake in the video. If it is true, then label "true"; otherwise, label "false". \\

Consider a YouTube comment from the math MIT OCW video below:\\
Playlist name: MIT 18.01 Single Variable Calculus, Fall 2006
Video name: Lec 35 | MIT 18.01 Single Variable Calculus, Fall 2007\\
Comment: can't L'Hopital's rule be explained geometricly? what about the functions curves' tangancy ?\\
Task: Does the comment ask a specific mathematical question and/or points out a mathematical mistake in the video?\\
Label: true\\

Consider a YouTube comment from the math MIT OCW video below:
Playlist name: MIT 18.06 Linear Algebra, Spring 2005\\
Video name: 14. Orthogonal Vectors and Subspaces\\
Comment: Just I have wondered. Are they student of MIT? Why are they so silent??????\\
Task: Does the comment ask a specific mathematical question and/or points out a mathematical mistake in the video?\\
Label: false\\

Consider a YouTube comment from the math MIT OCW video below:\\
Playlist name: MIT 18.100A Real Analysis, Fall 2020\\
Video name: Lecture 7: Convergent Sequences of Real Numbers\\
Comment: There is a mistake in lecture notes, example 71. Example in the lecture notes picks epsilon\_0=1\/2 and then proceeds with 1=|(-1)\^{}M-(-1)\^{}(M+1)|. This is wrong. Epsilon should be 1; and the expression with absolute values evaluates to 2. The lecture video is correct, the lecture notes are not.\\
Task: Does the comment ask a specific mathematical question and/or points out a mathematical mistake in the video?\\
Label: true\\

Consider a YouTube comment from the math MIT OCW video below:\\
Playlist name: \{playlistName\}\\
Video name: \{videoName\}\\
Comment: \{comment\}\\
Task: Does the comment ask a specific mathematical question and/or points out a mathematical mistake in the video?\\
Label
    \end{tcolorbox}
    \vspace{-1em}
    \caption{
    The k-shot prompt for the \confusion{} category.
    \label{fig:kshot_confusion_prompt}
    }
\end{figure*}
\begin{figure*}[t]
    \centering 
    \begin{tcolorbox}[
    kshot,
    title={\small \textbf{K-shot  prompting for \pedagogy{} category}},
    ]
    \small 
    Given a user comment on YouTube from a math MIT OCW video, your task is to label whether the comment explicitly mentions a pedagogical method, which includes but is not limited to the use of examples, applications, worked out problems, proofs, visualizations, elaboration, step-by-step explanation, reiteration, and analogies. If this is true, then label "true"; otherwise, label "false". \\ 

Consider a YouTube comment from the math MIT OCW video below: \\ 
Playlist name: MIT 18.06 Linear Algebra, Spring 2005 \\ 
Video name: 26. Complex Matrices; Fast Fourier Transform \\ 
Comment: He's just showing applications of linear algebra, not teaching them. That's why it seems "sloppy". You just can't teach Fourier Transform in 30 mins. \\ 
Task: Does the comment explicitly mention a pedagogical method? \\ 
Label: true \\ 

Consider a YouTube comment from the math MIT OCW video below: \\ 
Playlist name: MIT 18.06 Linear Algebra, Spring 2005 \\ 
Video name: 1. The Geometry of Linear Equations \\ 
Comment: This lecture plus 3blue1brown's videos are getting these concepts to stick for me. Thank you Prof. Strang!!! \\ 
Task: Does the comment explicitly mention a pedagogical method? \\ 
Label: false \\ 

Consider a YouTube comment from the math MIT OCW video below: \\ 
Playlist name: MIT 18.06 Linear Algebra, Spring 2005 \\ 
Video name: 9. Independence, Basis, and Dimension \\ 
Comment: His teaching style seems casual and intuitive. I go to a small public college and the course is much more formal and proof driven. These lectures are a great addition to (as well as a nice break from)  formal proofs. Thanks MIT! \\ 
Task: Does the comment explicitly mention a pedagogical method? \\ 
Label: true \\ 

Consider a YouTube comment from the math MIT OCW video below: \\ 
Playlist name: \{playlistName\}\\
Video name: \{videoName\}\\
Comment: \{comment\}\\
Task: Does the comment explicitly mention a pedagogical method? \\ 
Label: 
    \end{tcolorbox}
    \vspace{-1em}
    \caption{
    The k-shot prompt for the \pedagogy{} category.
    \label{fig:kshot_pedagogy_prompt}
    }
\end{figure*}
\begin{figure*}[t]
    \centering 
    \begin{tcolorbox}[
    kshot,
    title={\small \textbf{K-shot  prompting for \setup{} category}},
    ]
    \small 
    Given a user comment on YouTube from a math MIT OCW video, your task is to label whether the comment mentions the lecture’s physical teaching setup, which includes but is not limited to the chalk, board, microphone or audio-related aspects, and camera-related aspects (e.g., angle). If it is true, then label "true"; otherwise, label "false". \\ 

Consider a YouTube comment from the math MIT OCW video below: \\ 
Playlist name: MIT 18.100A Real Analysis, Fall 2020 \\ 
Video name: Lecture 1: Sets, Set Operations and Mathematical Induction \\ 
Comment: Thanks for posting this course, the instructor is great. If I may, there is only one request, in the future if the camera could move less frequently, the camera is following the instructor too closely, making me a bit dizzy. \\ 
Task: Does the comment mention the lecture’s physical teaching setup? \\ 
Label: true \\ 

Consider a YouTube comment from the math MIT OCW video below: \\ 
Playlist name: 6.041 Probabilistic Systems Analysis and Applied Probability \\ 
Video name: 5. Discrete Random Variables I \\ 
Comment:A ""random variable is a function in programming"".... mic drop! \\ 
Task: Does the comment mention the lecture’s physical teaching setup? \\ 
Label: false \\ 

Consider a YouTube comment from the math MIT OCW video below: \\ 
Playlist name: MIT 18.01 Single Variable Calculus, Fall 2006 \\ 
Video name: Lec 30 | MIT 18.01 Single Variable Calculus, Fall 2007 \\ 
Comment: The mic noise and hiss is distracting in this lecture, I hope someone could fix it .. \\ 
Task: Does the comment mention the lecture’s physical teaching setup? \\ 
Label: true \\ 

Consider a YouTube comment from the math MIT OCW video below: \\ 
Playlist name: \{playlistName\}\\
Video name: \{videoName\}\\
Comment: \{comment\}\\
Task: Does the comment mention the lecture’s physical teaching setup? \\ 
Label:
    \end{tcolorbox}
    \vspace{-1em}
    \caption{
    The k-shot prompt for the \setup{} category.
    \label{fig:kshot_setup_prompt}
    }
\end{figure*}
\begin{figure*}[t]
    \centering 
    \begin{tcolorbox}[
    kshot,
    title={\small \textbf{K-shot  prompting for \personal{} category}},
    ]
    \small 
    Given a user comment on YouTube from a math MIT OCW video, your task is to label whether the comment mentions the user’s personal experience learning or teaching math on their own outside of watching this lecture/series. If it is true, then label "true"; otherwise, label "false". \\ 

Consider a YouTube comment from the math MIT OCW video below: \\ 
Playlist name: MIT 18.06 Linear Algebra, Spring 2005 \\ 
Video name: 1. The Geometry of Linear Equations \\ 
Comment: Amazing! I like linear algebra a lot, I already had this class in college, I keep reading about it and ... I didn't even notice the passing of 40 minutes of the first class you! No wonder MIT is a world reference! \\ 
Task: Does the comment mention the user's personal experience learning or teaching math on their own outside of watching this lecture/series? \\ 
Label: true \\ 

Consider a YouTube comment from the math MIT OCW video below: \\ 
Playlist name: 6.041 Probabilistic Systems Analysis and Applied Probability \\ 
Video name: 14. Poisson Process I \\ 
Comment: I am having a hard time making sense of the notation at 11:22.  I believe the notation should be the conditional probability P(k|t) rather than P(k,t). I interpreted the latter to be the joint probability and if it is the case, the summation over all k of P(k,t) given a fixed t could not be equal to 1. Anyone, please help knock some sense to my head! \\ 
Task: Does the comment mention the user's personal experience learning or teaching math on their own outside of watching this lecture/series? \\ 
Label: false \\ 

Consider a YouTube comment from the math MIT OCW video below: \\ 
Playlist name: MIT 18.06 Linear Algebra, Spring 2005 \\ 
Video name: 21. Eigenvalues and Eigenvectors \\ 
Comment: Wish this guy taught me Math 293 and 294 at Cornell. My guy could barely speak English, let alone explain what we were trying to accomplish. I understood that if we wanted eigenvectors perpendicular to x we'd get lift relative to flow...but this guy would have made the math a bit simpler. \\ 
Task: Does the comment mention the user's personal experience learning or teaching math on their own outside of watching this lecture/series? \\ 
Label: true \\ 

Consider a YouTube comment from the math MIT OCW video below: \\ 
Playlist name: \{playlistName\}\\
Video name: \{videoName\}\\
Comment: \{comment\}\\
Task: Does the comment mention the user's personal experience learning or teaching math on their own outside of watching this lecture/series? \\ 
Label:
    \end{tcolorbox}
    \vspace{-1em}
    \caption{
    The k-shot prompt for the \personal{} category.
    \label{fig:kshot_personal_prompt}
    }
\end{figure*}
\begin{figure*}[t]
    \centering 
    \begin{tcolorbox}[
    kshot,
    title={\small \textbf{K-shot  prompting for \clarification{} category}},
    ]
    \small 
    Given a user comment on YouTube from a math MIT OCW video, your task is to label whether the comment clarifies someone's *math-related* misunderstanding or elaborates content from the video, and the comment includes an `@` that is immediately followed by a username. If this is true, then label "true"; otherwise, label "false". \\

Consider a YouTube comment from the math MIT OCW video below:\\
Playlist name: MIT 18.01 Single Variable Calculus, Fall 2006\\
Video name: Lec 3 | MIT 18.01 Single Variable Calculus, Fall 2007\\
Comment: @[USERNAME] it's the math dragon theorem\\
Task: Does the comment clarify someone's *math-related* misunderstanding or elaborate content from the video?\\
Label: true\\

Consider a YouTube comment from the math MIT OCW video below:\\
Playlist name: MIT 18.02 Multivariable Calculus, Fall 2007\\
Video name: Lec 23: Flux; normal form of Green's theorem | MIT 18.02 Multivariable Calculus, Fall 2007\\
Comment: 30:00, the way to remember it is that the work is a straightforward dot product of F with <dx, dy>, M goes with x and N goes with y and we add, and the flux is a dot product of F with the same vector rotated pi/2 so N goes with x and a minus sign with few choices left for M. Auroux missed a nice opportunity at the beginning to clarify the sign convention for flux by foreshadowing the result for closed curves with + being from the inside, out. I'm not faulting anyone, I couldn't give a lecture on this and keep possession of both my hands when erasing blackboards operated by hazardous machines. If he loses his hands, he'll never erase anything again. Be careful out there, Denis, we don't want to lose a great teacher.
Task: Does the comment clarify someone's *math-related* misunderstanding or elaborate content from the video?\\
Label: false\\

Consider a YouTube comment from the math MIT OCW video below:\\
Playlist name: MIT 18.01 Single Variable Calculus, Fall 2006\\
Video name: Lec 22 | MIT 18.01 Single Variable Calculus, Fall 2007\\
Comment: @[USERNAME] Actually, if a constant k=1\/1m is used, then in the final formula for V you will end up with subtracting m\^{} 1 from m\^{} 2 which is apparently not correct.\\
Task: Does the comment clarify someone's *math-related* misunderstanding or elaborate content from the video?\\
Label: true\\

Consider a YouTube comment from the math MIT OCW video below:\\
Playlist name: \{playlistName\}\\
Video name: \{videoName\}\\
Comment: \{comment\}\\
Task: Does the comment clarify someone's *math-related* misunderstanding or elaborate content from the video?\\
Label: 
    \end{tcolorbox}
    \vspace{-1em}
    \caption{
    The k-shot prompt for the \clarification{} category.
    \label{fig:kshot_clarification_prompt}
    }
\end{figure*}
\begin{figure*}[t]
    \centering 
    \begin{tcolorbox}[
    kshot,
    title={\small \textbf{K-shot  prompting for \gratitude{} category}},
    ]
    \small 
    Given a user comment on YouTube from a math MIT OCW video, your task is to label whether the comment contains the word "thanks" or "thank". If it is true, then label "true"; otherwise, label "false". \\

Consider a YouTube comment from the math MIT OCW video below: \\
Playlist name: MIT 18.650 Statistics for Applications, Fall 2016 \\
Video name: 15. Regression (cont.) \\
Comment: Thank you for the lectures, could you please state what topics did Lectures 10 and 16 covered? So we can research them separately. \\
Task: Does the comment contains the word "thanks" or "thank"? \\
Label: true \\

Consider a YouTube comment from the math MIT OCW video below: \\
Playlist name: MIT 18.100A Real Analysis, Fall 2020 \\
Video name: Lecture 1: Sets, Set Operations and Mathematical Induction \\
Comment: "Keep up the good work:thumbs\_up::thumbs\_up: \\
Task: Does the comment contains the word "thanks" or "thank"? \\
Label: false \\

Consider a YouTube comment from the math MIT OCW video below: \\
Playlist name: MIT 18.01 Single Variable Calculus, Fall 2006 \\
Video name: Lec 2 | MIT 18.01 Single Variable Calculus, Fall 2007 \\
Comment: Thanks! I prepared my high school final exam from this lecture. This really helped me!! \\
Task: Does the comment contains the word "thanks" or "thank"? \\
Label: true \\

Consider a YouTube comment from the math MIT OCW video below: \\
Playlist name: \{playlistName\}\\
Video name: \{videoName\}\\
Comment: \{comment\}\\
Task: Does the comment contains the word "thanks" or "thank"? \\
Label:
    \end{tcolorbox}
    \vspace{-1em}
    \caption{
    The k-shot prompt for the \gratitude{} category.
    \label{fig:kshot_gratitude_prompt}
    }
\end{figure*}
\begin{figure*}[t]
    \centering 
    \begin{tcolorbox}[
    kshot,
    title={\small \textbf{K-shot  prompting for \nonenglish{} category}},
    ]
    \small 
    Given a user comment on YouTube from a math MIT OCW video, your task is to label whether the comment is in English. If it is true, then label "true"; otherwise, label "false". \\

Consider a YouTube comment from the math MIT OCW video below: \\
Playlist name: MIT 18.06 Linear Algebra, Spring 2005 \\
Video name: 1. The Geometry of Linear Equations \\
Comment: Amazing! I like linear algebra a lot, I already had this class in college, I keep reading about it and ... I didn't even notice the passing of 40 minutes of the first class you! No wonder MIT is a world reference! \\
Task: Is the comment in English? \\
Label: true \\

Consider a YouTube comment from the math MIT OCW video below: \\
Playlist name: MIT 18.01 Single Variable Calculus, Fall 2006 \\
Video name: Lec 35 | MIT 18.01 Single Variable Calculus, Fall 2007 \\
Comment: \begin{CJK*}{UTF8}{mj}
이게 계속 쓰지 말라던 로피탈이구나
\end{CJK*}\\
Task: Is the comment in English? \\
Label: false \\

Consider a YouTube comment from the math MIT OCW video below: \\
Playlist name: MIT 18.06 Linear Algebra, Spring 2005 \\
Video name: 21. Eigenvalues and Eigenvectors \\
Comment: Wish this guy taught me Math 293 and 294 at Cornell. My guy could barely speak English, let alone explain what we were trying to accomplish. I understood that if we wanted eigenvectors perpendicular to x we'd get lift relative to flow...but this guy would have made the math a bit simpler. \\
Task: Is the comment in English? \\
Label: true \\

Consider a YouTube comment from the math MIT OCW video below: \\
Playlist name: \{playlistName\}\\
Video name: \{videoName\}\\
Comment: \{comment\}\\
Task: Is the comment in English? \\
Label: \\
    \end{tcolorbox}
    \vspace{-1em}
    \caption{
    The k-shot prompt for the \nonenglish{} category.
    \label{fig:kshot_nonenglish_prompt}
    }
\end{figure*}
\begin{figure*}[t]
    \centering 
    \begin{tcolorbox}[
    kshot,
    title={\small \textbf{K-shot  prompting for \na{} category}},
    ]
    \small 
    Given a user comment on YouTube from a math MIT OCW video, your task is to label whether the comment expresses a joke or is a troll comment. If it is true, then label "true"; otherwise, label "false". \\

Consider a YouTube comment from the math MIT OCW video below: \\
Playlist name: MIT 18.02 Multivariable Calculus, Fall 2007 \\
Video name: Lec 1: Dot product | MIT 18.02 Multivariable Calculus, Fall 2007 \\
Comment: Watching this to make me feel better about college algebra. lol \\
Task: Does the comment expresses a joke or is the comment a troll comment? \\
Label: true \\

Consider a YouTube comment from the math MIT OCW video below: \\
Playlist name: MIT 18.06 Linear Algebra, Spring 2005 \\
Video name: 3. Multiplication and Inverse Matrices \\
Comment: oh sir thank you a lot !!!! \\
Task: Does the comment expresses a joke or is the comment a troll comment? \\
Label: false \\

Consider a YouTube comment from the math MIT OCW video below: \\
Playlist name: MIT 18.02 Multivariable Calculus, Fall 2007 \\
Video name: Lec 24: Simply connected regions; review | MIT 18.02 Multivariable Calculus, Fall 2007 \\
Comment: i couldnt resist xD \\
Task: Does the comment expresses a joke or is the comment a troll comment? \\
Label: true \\

Consider a YouTube comment from the math MIT OCW video below: \\
Playlist name: \{playlistName\}\\
Video name: \{videoName\}\\
Comment: \{comment\}\\
Task: Does the comment expresses a joke or is the comment a troll comment? \\
Label:
    \end{tcolorbox}
    \vspace{-1em}
    \caption{
    The k-shot prompt for the \na{} category.
    \label{fig:kshot_na_prompt}
    }
\end{figure*}

\subsection{k-shot reasoning prompts}

\begin{itemize}
    \item Figure~\ref{fig:kshotreasoning_general_prompt} is the k-shot reasoning prompt for \general.
    \item Figure~\ref{fig:kshotreasoning_confusion_prompt} is the k-shot reasoning prompt for \confusion.
    \item Figure~\ref{fig:kshotreasoning_pedagogy_prompt} is the k-shot reasoning prompt for \pedagogy.
    \item Figure~\ref{fig:kshotreasoning_setup_prompt} is the k-shot reasoning prompt for \setup.
    \item Figure~\ref{fig:kshotreasoning_personal_prompt} is the k-shot reasoning prompt for \personal.
    \item Figure~\ref{fig:kshotreasoning_clarification_prompt} is the k-shot reasoning prompt for \clarification.
    \item Figure~\ref{fig:kshotreasoning_gratitude_prompt} is the k-shot reasoning prompt for \gratitude.
    \item Figure~\ref{fig:kshotreasoning_nonenglish_prompt} is the k-shot reasoning prompt for \nonenglish.
    \item Figure~\ref{fig:kshotreasoning_na_prompt} is the k-shot reasoning prompt for \na.
\end{itemize}

\begin{figure*}[t]
    \centering 
    \begin{tcolorbox}[
    kshotreasoning,
    title={\small \textbf{K-shot reasoning prompting for \general{} category}},
    ]
    \small 
    Given a user comment on YouTube from a math MIT OCW video, your task is to explain (after "Explanation:") and label (after "Label:") whether the comment expresses a general sentiment/adjective about or expresses a *general/big-picture* opinion about the video's *content* and/or about the teaching/professional characteristics of the *instructor*. If it is true, then label "true"; otherwise, label "false". \\ 

Consider a YouTube comment from the math MIT OCW video below: \\ 
Playlist name: MIT 18.06 Linear Algebra, Spring 2005 \\ 
Video name: 34. Final Course Review \\ 
Comment: Absolutely well done and definitely keep it up!!! :thumbs\_up::thumbs\_up:\\:thumbs\_up::thumbs\_up::thumbs\_up::thumbs\_up::thumbs\_up::thumbs\_up:\\:thumbs\_up::thumbs\_up::thumbs\_up::thumbs\_up::thumbs\_up::thumbs\_up: \\ 
Task: Does the comment express a general opinion about the video's content and/or about the teaching/professional characteristics of the instructor? \\ 
Explanation: Let's go through the sentences one by one until we find one that meets the criterion. "Absolutely well done and definitely keep it up!!! :thumbs\_up::thumbs\_up::thumbs\_up::thumbs\_up::thumbs\_up::thumbs\_up::thumbs\_up::thumbs\_up:\\:thumbs\_up::thumbs\_up::thumbs\_up::thumbs\_up::thumbs\_up::thumbs\_up:" expresses a general opinion about the video (well done). Therefore, the label is true. \\ 
Label: true \\ 

Consider a YouTube comment from the math MIT OCW video below: \\ 
Playlist name: MIT 18.02 Multivariable Calculus, Fall 2007 \\ 
Video name: Lec 3: Matrices; inverse matrices | MIT 18.02 Multivariable Calculus, Fall 2007 \\ 
Comment: Ideally, do you learn multivariable calculus first or linear algebra? A lot of stuff here seems to be based on 18.06. \\ 
Task: Does the comment express a general opinion about the video's content and/or about the teaching/professional characteristics of the instructor? \\ 
Explanation: Let's go through the sentences one by one until we find one that meets the criterion. "Ideally, do you learn multivariable calculus first or linear algebra?" asks a math-related question, and does not express a general opinion about the content or teaching of the instructor. "A lot of stuff here seems to be based on 18.06." builds on the math-related question. Therefore, the label is false. \\ 
Label: false \\ 

Consider a YouTube comment from the math MIT OCW video below: \\ 
Playlist name: MIT 18.02 Multivariable Calculus, Fall 2007 \\ 
Video name: Lec 16: Double integrals | MIT 18.02 Multivariable Calculus, Fall 2007 \\ 
Comment: This video is very helpful, i appreciate the help. \\ 
Task: Does the comment express a general opinion about the video's content and/or about the teaching/professional characteristics of the instructor? \\ 
Explanation: Let's go through the sentences one by one until we find one that meets the criterion. "This video is very helpful, i appreciate the help." expresses a general opinion of the video (helpful). Therefore, the label is true. \\ 
Label: true \\ 

Consider a YouTube comment from the math MIT OCW video below: \\ 
Playlist name: \{playlistName\}\\
Video name: \{videoName\}\\
Comment: \{comment\}\\
Task: Does the comment express a general opinion about the video's content and/or about the teaching/professional characteristics of the instructor? \\ 
Explanation:
    \end{tcolorbox}
    \vspace{-1em}
    \caption{
    The k-shot reasoning prompt for the \general{} category.
    \label{fig:kshotreasoning_general_prompt}
    }
\end{figure*}
\begin{figure*}[t]
    \centering 
    \begin{tcolorbox}[
    kshotreasoning,
    title={\small \textbf{K-shot reasoning prompting for \confusion{} category}},
    ]
    \small 
    Given a user comment on YouTube from a math MIT OCW video, your task is to explain (after "Explanation:") and label (after "Label:") whether the comment asks a specific mathematical question and/or points out a mathematical mistake in the video. If it is true, then label "true"; otherwise, label "false". \\ 

Consider a YouTube comment from the math MIT OCW video below: \\ 
Playlist name: MIT 18.01 Single Variable Calculus, Fall 2006 \\ 
Video name: Lec 35 | MIT 18.01 Single Variable Calculus, Fall 2007 \\ 
Comment: can't L'Hopital's rule be explained geometricly? what about the functions curves' tangancy ? \\ 
Task: Does the comment ask a specific mathematical question and/or points out a mathematical mistake in the video? \\ 
Explanation: Let's go through the sentences one by one until we find one that meets the criterion. "can't L'Hopital's rule be explained geometricly?" asks a question about L'Hopital's rule which is an important mathematical concept in calculus. Therefore, the label is true.  \\ 
Label: true \\ 

Consider a YouTube comment from the math MIT OCW video below: \\ 
Playlist name: MIT 18.06 Linear Algebra, Spring 2005 \\ 
Video name: 14. Orthogonal Vectors and Subspaces \\ 
Comment: Just I have wondered. Are they student of MIT? Why are they so silent?????? \\ 
Task: Does the comment ask a specific mathematical question and/or points out a mathematical mistake in the video? \\ 
Explanation: Let's go through the sentences one by one until we find one that meets the criterion. "Just I have wondered." is not a question and does not point out a mistake in the video. "Are they student of MIT? Why are they so silent??????" are questions, but it is not related to mathematics. Therefore, the label is false. \\ 
Label: false \\ 

Consider a YouTube comment from the math MIT OCW video below: \\ 
Playlist name: MIT 18.100A Real Analysis, Fall 2020 \\ 
Video name: Lecture 7: Convergent Sequences of Real Numbers \\ 
Comment: There is a mistake in lecture notes, example 71. Example in the lecture notes picks epsilon\_0=1/2 and then proceeds with 1=|(-1)\^{}M-(-1)\^{}(M+1)|. This is wrong. Epsilon should be 1; and the expression with absolute values evaluates to 2. The lecture video is correct, the lecture notes are not. \\ 
Task: Does the comment ask a specific mathematical question and/or points out a mathematical mistake in the video?
Explanation: Let's go through the sentences one by one until we find one that meets the criterion. "There is a mistake in lecture notes, example 71." points out a mistake in the lecture notes. Therefore, the label is true. \\ 
Label: true \\ 

Consider a YouTube comment from the math MIT OCW video below: \\ 
Playlist name: \{playlistName\}\\
Video name: \{videoName\}\\
Comment: \{comment\}\\
Task: Does the comment ask a specific mathematical question and/or points out a mathematical mistake in the video? \\ 
Explanation:
    \end{tcolorbox}
    \vspace{-1em}
    \caption{
    The k-shot reasoning prompt for the \confusion{} category.
    \label{fig:kshotreasoning_confusion_prompt}
    }
\end{figure*}
\begin{figure*}[t]
    \centering 
    \begin{tcolorbox}[
    kshotreasoning,
    title={\small \textbf{K-shot reasoning prompting for \pedagogy{} category}},
    ]
    \small 
    Given a user comment on YouTube from a math MIT OCW video, your task is to explain (after "Explanation:") and label (after "Label:") whether the comment explicitly mentions a pedagogical method, which includes but is not limited to the use of examples, applications, worked out problems, proofs, visualizations, elaboration, step-by-step explanation, reiteration, and analogies. If this is true, then label "true"; otherwise, label "false". \\ 

Consider a YouTube comment from the math MIT OCW video below: \\ 
Playlist name: MIT 18.06 Linear Algebra, Spring 2005 \\ 
Video name: 26. Complex Matrices; Fast Fourier Transform \\ 
Comment: He's just showing applications of linear algebra, not teaching them. That's why it seems "sloppy". You just can't teach Fourier Transform in 30 mins. \\ 
Task: Does the comment explicitly mention a pedagogical method? \\ 
Explanation: Let's go through the sentences one by one until we find one that meets the criterion. "He's just showing applications of linear algebra, not teaching them." mentions the teacher is using applications. Applications are a pedagogical methods. Therefore, the label is true.  \\ 
Label: true \\ 

Consider a YouTube comment from the math MIT OCW video below: \\ 
Playlist name: MIT 18.06 Linear Algebra, Spring 2005 \\ 
Video name: 1. The Geometry of Linear Equations \\ 
Comment: This lecture plus 3blue1brown's videos are getting these concepts to stick for me. Thank you Prof. Strang!!!
Task: Does the comment explicitly mention a pedagogical method? \\ 
Explanation: Let's go through the sentences one by one until we find one that meets the criterion. "This lecture plus 3blue1brown's videos are getting these concepts to stick for me." communicates that the video is helpful, but it does not mention any pedagogical method that makes the video helpful. "Thank you Prof. Strang!!!" does not mention any pedagogical method. Therefore, the label is false.  \\ 
Label: false \\ 

Consider a YouTube comment from the math MIT OCW video below: \\ 
Playlist name: MIT 18.06 Linear Algebra, Spring 2005 \\ 
Video name: 9. Independence, Basis, and Dimension \\ 
Comment: His teaching style seems casual and intuitive. I go to a small public college and the course is much more formal and proof driven. These lectures are a great addition to (as well as a nice break from)  formal proofs. Thanks MIT! \\ 
Task: Does the comment explicitly mention a pedagogical method? \\ 
Explanation: Let's go through the sentences one by one until we find one that meets the criterion. "His teaching style seems casual and intuitive." describes the teaching style, but does not mention what methods the instructor uses to enable for a casual and intuitive style. "I go to a small public college and the course is much more formal and proof driven." mentions the proofs from their previous course, which is a pedagogical method. Therefore, the label is true. \\ 
Label: true \\ 

Consider a YouTube comment from the math MIT OCW video below: \\ 
Playlist name: \{playlistName\}\\
Video name: \{videoName\}\\
Comment: \{comment\}\\
Task: Does the comment explicitly mention a pedagogical method? \\ 
Explanation: 
    \end{tcolorbox}
    \vspace{-1em}
    \caption{
    The k-shot reasoning prompt for the \pedagogy{} category.
    \label{fig:kshotreasoning_pedagogy_prompt}
    }
\end{figure*}
\begin{figure*}[t]
    \centering 
    \begin{tcolorbox}[
    kshotreasoning,
    title={\small \textbf{K-shot reasoning prompting for \setup{} category}},
    ]
    \small 
    Given a user comment on YouTube from a math MIT OCW video, your task is to explain (after "Explanation:") and label (after "Label:") whether the comment mentions the lecture’s physical teaching setup, which includes but is not limited to the chalk, board, microphone or audio-related aspects, and camera-related aspects (e.g., angle). If it is true, then label "true"; otherwise, label "false".\\

Consider a YouTube comment from the math MIT OCW video below:\\
Playlist name: MIT 18.100A Real Analysis, Fall 2020\\
Video name: Lecture 1: Sets, Set Operations and Mathematical Induction\\
Comment: Thanks for posting this course, the instructor is great. If I may, there is only one request, in the future if the camera could move less frequently, the camera is following the instructor too closely, making me a bit dizzy.\\
Task: Does the comment mention the lecture’s physical teaching setup?\\
Explanation: Let's go through the sentences one by one until we find one that meets the criterion. "Thanks for posting this course, the instructor is great." does not mention the lecture's physical teaching setup. "If I may, there is only one request, in the future if the camera could move less frequently, the camera is following the instructor too closely, making me a bit dizzy." mentions the camera, which is a part of the lecture's physical setup. Therefore, the label is true.\\
Label: true\\

Consider a YouTube comment from the math MIT OCW video below:\\
Playlist name: 6.041 Probabilistic Systems Analysis and Applied Probability\\
Video name: 5. Discrete Random Variables I\\
Comment: A ""random variable is a function in programming"".... mic drop!\\
Task: Does the comment mention the lecture’s physical teaching setup?\\
Explanation: Let's go through the sentences one by one until we find one that meets the criterion. "A ""random variable is a function in programming"".... mic drop!" mentions a mic, but is used figuratively in this context. Therefore, the label is false. \\
Label: false\\

Consider a YouTube comment from the math MIT OCW video below:\\
Playlist name: MIT 18.01 Single Variable Calculus, Fall 2006\\
Video name: Lec 30 | MIT 18.01 Single Variable Calculus, Fall 2007\\
Comment: The mic noise and hiss is distracting in this lecture, I hope someone could fix it ..\\
Task: Does the comment mention the lecture’s physical teaching setup?\\
Explanation: Let's go through the sentences one by one until we find one that meets the criterion. "The mic noise and hiss is distracting in this lecture, I hope someone could fix it .." mentions the mic hissing, which is part of the physical teaching setup. Therefore, the label is true. \\
Label: true\\

Consider a YouTube comment from the math MIT OCW video below:\\
Playlist name: \{playlistName\}\\
Video name: \{videoName\}\\
Comment: \{comment\}\\
Task: Does the comment mention the lecture’s physical teaching setup?\\
Explanation:
    \end{tcolorbox}
    \vspace{-1em}
    \caption{
    The k-shot reasoning prompt for the \setup{} category.
    \label{fig:kshotreasoning_setup_prompt}
    }
\end{figure*}
\begin{figure*}[t]
    \centering 
    \begin{tcolorbox}[
    kshotreasoning,
    title={\small \textbf{K-shot reasoning prompting for \personal{} category}},
    ]
    \small 
    Given a user comment on YouTube from a math MIT OCW video, your task is to explain (after "Explanation:") and label (after "Label:") whether the comment mentions the user’s personal experience learning or teaching math on their own outside of watching this lecture/series. If it is true, then label "true"; otherwise, label "false". \\ 

Consider a YouTube comment from the math MIT OCW video below: \\ 
Playlist name: MIT 18.06 Linear Algebra, Spring 2005 \\ 
Video name: 1. The Geometry of Linear Equations \\ 
Comment: Amazing! I like linear algebra a lot, I already had this class in college, I keep reading about it and ... I didn't even notice the passing of 40 minutes of the first class you! No wonder MIT is a world reference! \\ 
Task: Does the comment mention the user's personal experience learning or teaching math on their own outside of watching this lecture/series?  \\ 
Explanation: Let's go through the sentences one by one until we find one that mentions the user’s personal experience. "Amazing!" expresses the user's opinion about the content, but does not mention their personal experience outside of this lecture. "I like linear algebra a lot, I already had this class in college, I keep reading about it and ... I didn't even notice the passing of 40 minutes of the first class you!" mentions taking this class in college, which is a personal experience for this user outside of watching this lecture or series. Therefore, the label is true. \\ 
Label: true \\ 

Consider a YouTube comment from the math MIT OCW video below: \\ 
Playlist name: 6.041 Probabilistic Systems Analysis and Applied Probability \\ 
Video name: 14. Poisson Process I \\ 
Comment: I am having a hard time making sense of the notation at 11:22.  I believe the notation should be the conditional probability P(k|t) rather than P(k,t). I interpreted the latter to be the joint probability and if it is the case, the summation over all k of P(k,t) given a fixed t could not be equal to 1. Anyone, please help knock some sense to my head! \\ 
Task: Does the comment mention the user's personal experience learning or teaching math on their own outside of watching this lecture/series? \\ 
Explanation: Let's go through the sentences one by one until we find one that mentions the user’s personal experience. "I am having a hard time making sense of the notation at 11:22." expresses the user's confusion with the lecture content, but not an experience outside of watching this lecture or series. "I believe the notation should be the conditional probability P(k|t) rather than P(k,t)." elaborates what the user is confused about with the lecture, but not a personal experience outside of the lecture or series. "I interpreted the latter to be the joint probability and if it is the case, the summation over all k of P(k,t) given a fixed t could not be equal to 1." elaborates what the user misunderstood, but does not communicate a personal experience outside of watching thie lecture or series. "Anyone, please help knock some sense to my head!" requests for help from others, but does not talk about a personal experience outside of this lecture or series. Therefore, the label is false.  \\ 
Label: false \\ 

Consider a YouTube comment from the math MIT OCW video below: \\ 
Playlist name: MIT 18.06 Linear Algebra, Spring 2005 \\ 
Video name: 21. Eigenvalues and Eigenvectors \\ 
Comment: Wish this guy taught me Math 293 and 294 at Cornell. My guy could barely speak English, let alone explain what we were trying to accomplish. I understood that if we wanted eigenvectors perpendicular to x we'd get lift relative to flow...but this guy would have made the math a bit simpler. \\ 
Task: Does the comment mention the user's personal experience learning or teaching math on their own outside of watching this lecture/series? \\ 
Explanation: Let's go through the sentences one by one until we find one that mentions the user’s personal experience. "Wish this guy taught me Math 293 and 294 at Cornell." mentions the user's own math classes at a different university. This is a personal experience related to learning outside of this video and lecture series. Therefore, the label is true. \\ 
Label: true \\ 

Consider a YouTube comment from the math MIT OCW video below: \\ 
Playlist name: \{playlistName\}\\
Video name: \{videoName\}\\
Comment: \{comment\}\\
Task: Does the comment mention the user's personal experience learning or teaching math on their own outside of watching this lecture/series? \\ 
Explanation:
    \end{tcolorbox}
    \vspace{-1em}
    \caption{
    The k-shot reasoning prompt for the \personal{} category.
    \label{fig:kshotreasoning_personal_prompt}
    }
\end{figure*}
\begin{figure*}[t]
    \centering 
    \begin{tcolorbox}[
    kshotreasoning,
    title={\small \textbf{K-shot reasoning prompting for \clarification{} category}},
    ]
    \small 
    Given a user comment on YouTube from a math MIT OCW video, your task is to explain (after "Explanation:") and label (after "Label:") whether the comment clarifies someone's *math-related* misunderstanding or elaborates content from the video, and the comment includes an `@` that is immediately followed by a username. If this is true, then label "true"; otherwise, label "false". \\ 

Consider a YouTube comment from the math MIT OCW video below: \\ 
Playlist name: MIT 18.01 Single Variable Calculus, Fall 2006 \\ 
Video name: Lec 3 | MIT 18.01 Single Variable Calculus, Fall 2007 \\ 
Comment: @[USERNAME] it's the math dragon theorem \\ 
Task: Does the comment clarify someone's *math-related* misunderstanding or elaborate content from the video? \\ 
Explanation: Let's go through the sentences one by one until we find one that meets the criterion. "@[USERNAME] it's the math dragon theorem" tags another user, and seems to respond to a question from this user. Responding to a question is a form of clarification. Therefore, the label is true. \\ 
Label: true \\ 

Consider a YouTube comment from the math MIT OCW video below: \\ 
Playlist name: MIT 18.02 Multivariable Calculus, Fall 2007 \\ 
Video name: Lec 23: Flux; normal form of Green's theorem | MIT 18.02 Multivariable Calculus, Fall 2007 \\ 
Comment: 30:00, the way to remember it is that the work is a straightforward dot product of F with <dx, dy>, M goes with x and N goes with y and we add, and the flux is a dot product of F with the same vector rotated pi/2 so N goes with x and a minus sign with few choices left for M. Auroux missed a nice opportunity at the beginning to clarify the sign convention for flux by foreshadowing the result for closed curves with + being from the inside, out. I'm not faulting anyone, I couldn't give a lecture on this and keep possession of both my hands when erasing blackboards operated by hazardous machines. If he loses his hands, he'll never erase anything again. Be careful out there, Denis, we don't want to lose a great teacher. \\ 
Task: Does the comment clarify someone's *math-related* misunderstanding or elaborate content from the video? \\ 
Explanation: Let's go through the sentences one by one until we find one that meets the criterion. "30:00, the way to remember it is that the work is a straightforward dot product of F with <dx, dy>, M goes with x and N goes with y and we add, and the flux is a dot product of F with the same vector rotated pi/2 so N goes with x and a minus sign with few choices left for M." does not contain any @ symbol. "Auroux missed a nice opportunity at the beginning to clarify the sign convention for flux by foreshadowing the result for closed curves with + being from the inside, out." also does not contain the @ symbol. "Also  I'm not faulting anyone, I couldn't give a lecture on this and keep possession of both my hands when erasing blackboards operated by hazardous machines." also does not contain the @ symbol. "If he loses his hands, he'll never erase anything again. Be careful out there, Denis, we don't want to lose a great teacher." also does not contain the @ symbol. Therefore the label is false. \\  
Label: false \\ 

Consider a YouTube comment from the math MIT OCW video below: \\ 
Playlist name: MIT 18.01 Single Variable Calculus, Fall 2006 \\ 
Video name: Lec 22 | MIT 18.01 Single Variable Calculus, Fall 2007 \\ 
Comment: @[USERNAME] Actually, if a constant k=1\/1m is used, then in the final formula for V you will end up with subtracting m\^{}1 from m\^{}2 which is apparently not correct. \\ 
Task: Does the comment clarify someone's *math-related* misunderstanding or elaborate content from the video? \\ 
Explanation: Let's go through the sentences one by one until we find one that meets the criterion. "@[USERNAME] Actually, if a constant k=1\/1m is used, then in the final formula for V you will end up with subtracting m\^{}1 from m\^{}2 which is apparently not correct." contains the @ symbol and seems to correct the other user's understanding of the math formula. Correcting someone's understanding of the math formula is a form of clarification. Therefore, the label is true. \\ 
Label: true \\ 

Consider a YouTube comment from the math MIT OCW video below: \\ 
Playlist name: \{playlistName\}\\
Video name: \{videoName\}\\
Comment: \{comment\}\\
Task: Does the comment clarify someone's *math-related* misunderstanding or elaborate content from the video? \\ 
Explanation:
    \end{tcolorbox}
    \vspace{-1em}
    \caption{
    The k-shot reasoning prompt for the \clarification{} category.
    \label{fig:kshotreasoning_clarification_prompt}
    }
\end{figure*}
\begin{figure*}[t]
    \centering 
    \begin{tcolorbox}[
    kshotreasoning,
    title={\small \textbf{K-shot reasoning prompting for \gratitude{} category}},
    ]
    \small 
    Given a user comment on YouTube from a math MIT OCW video, your task is to explain (after "Explanation:") and label (after "Label:") whether the comment contains the word "thanks" or "thank". If it is true, then label "true"; otherwise, label "false". \\ 

Consider a YouTube comment from the math MIT OCW video below: \\ 
Playlist name: MIT 18.650 Statistics for Applications, Fall 2016 \\ 
Video name: 15. Regression (cont.) \\ 
Comment: Thank you for the lectures, could you please state what topics did Lectures 10 and 16 covered? So we can research them separately. \\ 
Task: Does the comment contains the word "thanks" or "thank"? \\ 
Explanation: Let's go through the sentences one by one until we find one that meets the criterion. "Thank you for the lectures, could you please state what topics did Lectures 10 and 16 covered?" contains one of the expressions. Therefore, the label is true.  \\ 
Label: true \\ 

Consider a YouTube comment from the math MIT OCW video below: \\ 
Playlist name: MIT 18.100A Real Analysis, Fall 2020 \\ 
Video name: Lecture 1: Sets, Set Operations and Mathematical Induction \\ 
Comment: Keep up the good work:thumbs\_up::thumbs\_up: \\ 
Task: Does the comment contains the word "thanks" or "thank"? \\ 
Explanation: Let's go through the sentences one by one until we find one that meets the criterion. "Keep up the good work:thumbs\_up::thumbs\_up:" does not contain any of the expressions. Therefore, the label is false. \\ 
Label: false \\ 

Consider a YouTube comment from the math MIT OCW video below: \\ 
Playlist name: MIT 18.01 Single Variable Calculus, Fall 2006 \\ 
Video name: Lec 2 | MIT 18.01 Single Variable Calculus, Fall 2007 \\ 
Comment: Thanks! I prepared my high school final exam from this lecture. This really helped me!! \\ 
Task: Does the comment contains the word "thanks" or "thank"? \\ 
Explanation: Let's go through the sentences one by one until we find one that meets the criterion. "Thanks!" contains one of the expressions. Therefore, the label is true. \\ 
Label: true \\ 

Consider a YouTube comment from the math MIT OCW video below: \\ 
Playlist name: \{playlistName\}\\
Video name: \{videoName\}\\
Comment: \{comment\}\\
Task: Does the comment contains the word "thanks" or "thank"? \\ 
Explanation: 
    \end{tcolorbox}
    \vspace{-1em}
    \caption{
    The k-shot reasoning prompt for the \gratitude{} category.
    \label{fig:kshotreasoning_gratitude_prompt}
    }
\end{figure*}
\begin{figure*}[t]
    \centering 
    \begin{tcolorbox}[
    kshotreasoning,
    title={\small \textbf{K-shot reasoning prompting for \nonenglish{} category}},
    ]
    \small 
    Given a user comment on YouTube from a math MIT OCW video, your task is to explain (after "Explanation:") and label (after "Label:") whether the comment is in English. If it is true, then label "true"; otherwise, label "false". \\ 

Consider a YouTube comment from the math MIT OCW video below: \\ 
Playlist name: MIT 18.06 Linear Algebra, Spring 2005 \\ 
Video name: 1. The Geometry of Linear Equations \\ 
Comment: Amazing! I like linear algebra a lot, I already had this class in college, I keep reading about it and ... I didn't even notice the passing of 40 minutes of the first class you! No wonder MIT is a world reference! \\ 
Task: Is the comment in English? \\ 
Explanation: Let's go through the sentences one by one until we find a sentence that is not in English. "Amazing!" is in English. "I like linear algebra a lot, I already had this class in college, I keep reading about it and ... I didn't even notice the passing of 40 minutes of the first class you!" is in English. "No wonder MIT is a world reference!" is also in English. The entire comment is in English. Therefore, the label is true. \\ 
Label: true \\ 

Consider a YouTube comment from the math MIT OCW video below: \\ 
Playlist name: MIT 18.01 Single Variable Calculus, Fall 2006 \\ 
Video name: Lec 35 | MIT 18.01 Single Variable Calculus, Fall 2007 \\ 
Comment: \begin{CJK*}{UTF8}{mj}
이게 계속 쓰지 말라던 로피탈이구나
\end{CJK*}\\
Task: Is the comment in English? \\ 
Explanation: Let's go through the sentences one by one until we find a sentence that is not in English. \begin{CJK*}{UTF8}{mj}
이게 계속 쓰지 말라던 로피탈이구나
\end{CJK*} is not in English. Therefore, the label is false.  \\ 
Label: false \\ 

Consider a YouTube comment from the math MIT OCW video below: \\ 
Playlist name: MIT 18.06 Linear Algebra, Spring 2005 \\ 
Video name: 21. Eigenvalues and Eigenvectors \\ 
Comment: Wish this guy taught me Math 293 and 294 at Cornell. My guy could barely speak English, let alone explain what we were trying to accomplish. I understood that if we wanted eigenvectors perpendicular to x we'd get lift relative to flow...but this guy would have made the math a bit simpler. \\ 
Task: Is the comment in English? \\ 
Explanation: Let's go through the sentences one by one until we find a sentence that is not in English. "Wish this guy taught me Math 293 and 294 at Cornell." is in English. "My guy could barely speak English, let alone explain what we were trying to accomplish." is in English. "I understood that if we wanted eigenvectors perpendicular to x we'd get lift relative to flow...but this guy would have made the math a bit simpler." is also in English. The entire comment is in English. Therefore, the label is true. \\ 
Label: true \\ 

Consider a YouTube comment from the math MIT OCW video below: \\ 
Playlist name: \{playlistName\}\\
Video name: \{videoName\}\\
Comment: \{comment\}\\
Task: Is the comment in English? \\ 
Explanation: 
    \end{tcolorbox}
    \vspace{-1em}
    \caption{
    The k-shot reasoning prompt for the \nonenglish{} category.
    \label{fig:kshotreasoning_nonenglish_prompt}
    }
\end{figure*}
\begin{figure*}[t]
    \centering 
    \begin{tcolorbox}[
    kshotreasoning,
    title={\small \textbf{K-shot reasoning prompting for \na{} category}},
    ]
    \small 
    Given a user comment on YouTube from a math MIT OCW video, your task is to explain (after "Explanation:") and label (after "Label:") whether the comment expresses a joke or is a troll comment. If it is true, then label "true"; otherwise, label "false". \\ 

Consider a YouTube comment from the math MIT OCW video below: \\ 
Playlist name: MIT 18.02 Multivariable Calculus, Fall 2007 \\ 
Video name: Lec 1: Dot product | MIT 18.02 Multivariable Calculus, Fall 2007 \\ 
Comment: Watching this to make me feel better about college algebra. lol \\ 
Task: Does the comment expresses a joke or is the comment a troll comment? \\ 
Explanation: Let's go through the sentences one by one until we find one that meets the criterion. "Watching this to make me feel better about college algebra." does not seem to express a joke. "lol" expresses a joking tone to the comment. Therefore, the label is true.  \\ 
Label: true \\ 

Consider a YouTube comment from the math MIT OCW video below: \\ 
Playlist name: MIT 18.06 Linear Algebra, Spring 2005 \\ 
Video name: 3. Multiplication and Inverse Matrices \\ 
Comment: oh sir thank you a lot !!!! \\ 
Task: Does the comment expresses a joke or is the comment a troll comment? \\ 
Explanation: Let's go through the sentences one by one until we find one that meets the criterion. "oh sir thank you a lot !!!!" does not express a joke or troll comment. Therefore, the label is false. \\ 
Label: false \\ 

Consider a YouTube comment from the math MIT OCW video below: \\ 
Playlist name: MIT 18.02 Multivariable Calculus, Fall 2007 \\ 
Video name: Lec 24: Simply connected regions; review | MIT 18.02 Multivariable Calculus, Fall 2007 \\ 
Comment: i couldnt resist xD \\ 
Task: Does the comment expresses a joke or is the comment a troll comment? \\ 
Explanation: Let's go through the sentences one by one until we find one that meets the criterion. "i couldnt resist xD" ends in a joking emoji and expressing a joking tone. Therefore, the label is true. \\ 
Label: true \\ 

Consider a YouTube comment from the math MIT OCW video below: \\ 
Playlist name: \{playlistName\}\\
Video name: \{videoName\}\\
Comment: \{comment\}\\
Task: Does the comment expresses a joke or is the comment a troll comment? \\ 
Explanation:
    \end{tcolorbox}
    \vspace{-1em}
    \caption{
    The k-shot reasoning prompt for the \na{} category.
    \label{fig:kshotreasoning_na_prompt}
    }
\end{figure*}

\end{document}